\documentclass[10pt,twocolumn,letterpaper]{article}
\usepackage{times}
\usepackage{epsfig}
\usepackage{graphicx}
\usepackage{amsmath}
\usepackage{amssymb}
\usepackage{subcaption}
\usepackage{multirow}

\usepackage{caption}

\begin{document}
	
	\title{Time-Aware and View-Aware Video Rendering\\ for Unsupervised Representation Learning}
	
	\author{Shruti Vyas\\
		{\tt\small shruti@crcv.ucf.edu}
		\and
		Yogesh S Rawat\\
		{\tt\small yogesh@crcv.ucf.edu}
		\and 
		Mubarak Shah\\
		{\tt\small shah@crcv.ucf.edu}\\
		\and 
		Center for Research in Computer Vision\\
		University of Central Florida\\
	}
	
	\maketitle
	
	\begin{abstract}
		
		
		The recent success in deep learning has lead to various effective representation learning methods for videos. However, the current approaches for video representation require large amount of human labeled datasets for effective learning. We present an unsupervised representation learning framework to encode scene dynamics in videos captured from multiple viewpoints. The proposed framework has two main components: Representation Learning Network (RL-NET), which learns a representation with the help of Blending Network (BL-NET), and Video Rendering Network (VR-NET), which is used for video synthesis. The framework takes as input video clips from different viewpoints and time, learns an internal representation and uses this representation to render a video clip from an arbitrary given viewpoint and time. The ability of the proposed network to render video frames from arbitrary viewpoints and time enable it to learn a meaningful and robust representation of the scene dynamics. 
		We demonstrate the effectiveness of the proposed method in rendering view-aware as well as time-aware video clips on two different real-world  datasets including UCF-101 and NTU-RGB+D. To further validate the effectiveness of the learned representation, we use it for the task of view-invariant activity classification where we observe a significant improvement ($\sim 26\%$) in the performance on NTU-RGB+D dataset  compared to the existing state-of-the art methods.
	
	\end{abstract}

	\begin{figure}[t]
		\begin{center}
			\includegraphics[width=0.85\linewidth]{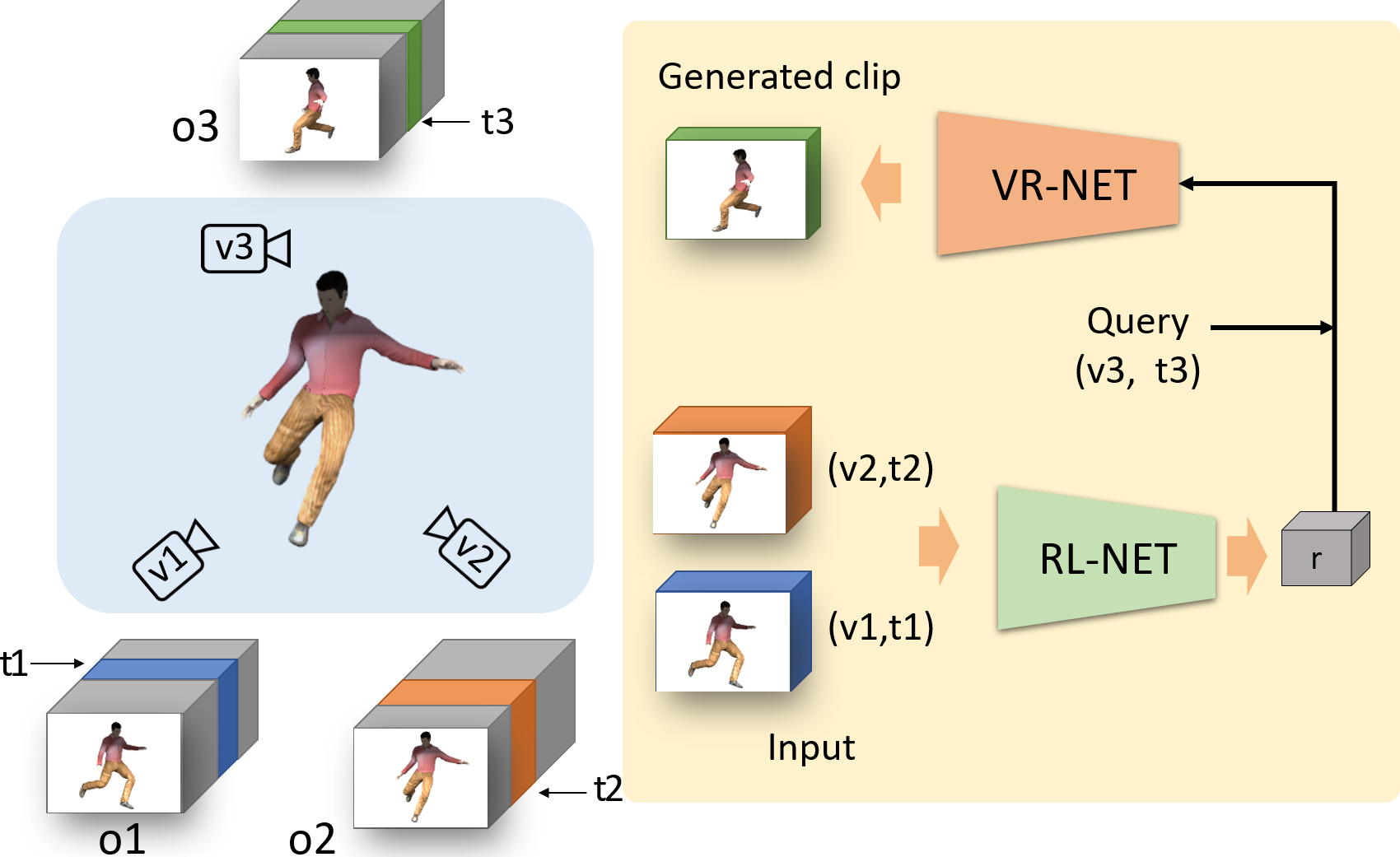}
		\end{center}
		\caption{\small An overview of the proposed video rendering framework. An activity is captured from different viewpoints (v1, v2, and v3) providing observations (o1, o2, and o3). Video clips from these viewpoints (v1 and v2) at arbitrary times (t1 and t2) are used to learn a scene and dynamics representation (r) for this activity, employing the  proposed RL-NET. The learned representation (r) is then used to render a video from an arbitrary query viewpoint (v3) and time (t3) using proposed VR-NET.}
		\label{fig1}
	\end{figure}
	
	\section{Introduction}
	In recent years, we have seen some success in the research on video synthesis. The proposed methods are mainly focused on future frame prediction \cite{mathieu2015deep, babaeizadeh2017stochastic, byeon2018contextvp}, future clip prediction \cite{vondrick2016generating, xiong2017learning, kratzwald2017towards}, or conditioned video generation \cite{tulyakov2017mocogan, baddar2017dynamics, spampinato2018vos, saito2017temporal}. The future frame/clip prediction methods can synthesize frames from near future in the video with some realism. The video generation works mainly leverage the Generative Adversarial Network \cite{goodfellow2014generative} for realistic video synthesis based on some conditioning. All of these methods tackle the problem of video synthesis from a single viewpoint and focus on the time dimension of the video. However, wouldn't be interesting if we can bring in the notion of viewpoint into video synthesis as well? This is what we address in this paper.
	
	The diversity in a scene and objects dynamics, along with camera motion, makes video synthesis of real-world scenarios a very challenging problem. Adding the notion of viewpoint makes it even more difficult, as the observed video of the same scene and dynamics will vary from one viewpoint to another. Historically, view-point in-variance has been a very active research area in computer vision, and is currently also important from the perspective of representation learning. There have been some work in view-aware {\it image} synthesis \cite{regmi2018cross, eslami2018neural}, where given images captured from different views one can synthesize an image from an unseen view.
	Most intriguing recent work in this area is by Eslami et. al\cite {eslami2018neural}, who proposed a simple neural network approach for rendering {\em images} from unseen views for {\em synthetic} data.
	However, to the best of our knowledge, the problem of view-aware {\em video} synthesis has not been addressed yet. Inspired by Eslami et. al\cite {eslami2018neural}, in this paper, we explore generalization of their idea to video and propose time-aware and view-aware video rendering for real-world videos instead of synthetic images.
	
	
	The proposed framework, shown in  Figure \ref{fig1},  takes multiple video clips from varying view-points and times as input, and renders a video from a given an arbitrary viewpoint and time. The learned representation has an understanding of temporal and view-point significance of the input videos. The ability of the network to render a video from any given viewpoint and time, enables the network to learn a robust view and time aware representation, which can be employed for view-invariant activity classification.
	
	We make the following contributions in this paper. We introduce a novel view and time-aware video rendering problem and propose an unsupervised representation learning framework to solve this. The proposed framework bears an understanding of time and view-point and therefore the learned representation can be used to generate videos from arbitrary given viewpoint and time. We also demonstrate the effectiveness of the proposed framework for future video prediction. We further validate the robustness of the learned representation for view-invariant action classification, where we observe a significant improvement ($\sim 26\%$) in the performance, when compared with the state-of-the art methods on NTU-RGB+D dataset using RGB modality.

	\section{Related Work}
	The research in image synthesis \cite{mirza2014conditional, gregor2015draw, wang2017high} has recently seen a great progress, and it is mainly attributed to the success of Generative Adversarial Networks (GAN) \cite{goodfellow2014generative}. This includes methods for generating high-resolution images \cite{ledig2017photo, wang2017high} along with image-to-image translation \cite{zhu2017unpaired, liu2017unsupervised, choi2017stargan}. However, video synthesis is still a landmark challenge and the research is in preliminary stage.
	
	Our work is closely related to the research in video synthesis, which explores future frame prediction \cite{mathieu2015deep, babaeizadeh2017stochastic, byeon2018contextvp}, future clip prediction \cite{xiong2017learning, kratzwald2017towards}, and conditioned video generation \cite{vondrick2016generating, tulyakov2017mocogan, baddar2017dynamics, spampinato2018vos, saito2017temporal}. In one of the early attempts on video prediction, the authors in \cite{mathieu2015deep} proposed a GAN based approach for next-frame prediction, where they explore different types of loss functions along with adversarial loss. Similarly, the authors in \cite{vondrick2016generating, baddar2017dynamics, xiong2017learning, kratzwald2017towards, tulyakov2017mocogan, saito2017temporal, spampinato2018vos} explored the GAN framework for video synthesis, where they focus on future frame prediction and conditional video generation. The authors in \cite{babaeizadeh2017stochastic} recently proposed a variational latent space learning framework for video synthesis, and similarly the authors in \cite{byeon2018contextvp} also explored a recurrent approach for next frame prediction. {\em Our work is distinctly different from these approaches as we have a notion of viewpoint, which has not been addressed earlier}. Apart from this, our proposed method is time-aware, and therefore given a single-view video it can also be used for future frame prediction.
	
	In addition, our work is also related to unsupervised video representation learning \cite{srivastava2015unsupervised, Wang_2015_ICCV, goyal2017nonparametric}. The research in unsupervised video representation learning mainly focuses on encoder-decoder \cite{kingma2013auto} kind of networks, where the decoder is used for reconstruction of the input video. The authors in \cite{srivastava2015unsupervised} proposed a recurrent network for both encoding and decoding which was based on LSTM. Similarly, the authors in \cite{goyal2017nonparametric} proposed to learn a latent distribution instead of encoding for video representation. Our approach resembles to these methods as we  also learn a representation. However, instead of learning a representation for a single video, we learn a representation for the whole scene and its dynamics  captured from different viewpoints. Also, instead of trying to generate the input video, we aim to generate a video from a different viewpoint and time.
	
	Cross-view synthesis of data is an interesting problem, which can have multiple applications including view-invariant representation learning. There are some existing works focusing on this problem for image synthesis \cite{regmi2018cross, eslami2018neural}. In \cite{rahmani20163d}, the authors proposed a GAN based approach, where they perform a image-to-image translation from aerial to ground view images. In \cite{eslami2018neural}, the authors proposed a scene representation learning framework where a representation is learned for a scene and a view of the same scene is generated from a different viewpoint. However, there has been no work in cross-view video synthesis. Our work is inspired from \cite{eslami2018neural}, which was mainly focused on synthetic {\em images} and we generalize it to {\em real-world videos}. Apart from this, we also propose the notion of {\em time awareness} in the viewpoint which is intuitive from the perspective of videos. 
	
	We demonstrate the effectiveness of the learned representation for view-invariant activity classification. It is an active research topic in computer vision community \cite{shahroudy2018deep} and most of the existing works are focused on multiple modalities, such as depth \cite{10.1007/978-3-319-46448-0_32} and pose \cite{shahroudy2016ntu}, besides RGB data \cite{luo2017unsupervised, li2018unsupervised}. We mainly focus on RGB modality and present a comparison with existing works.
	
	\begin{figure*}[t]
		\begin{center}
			\includegraphics[width=0.85\linewidth]{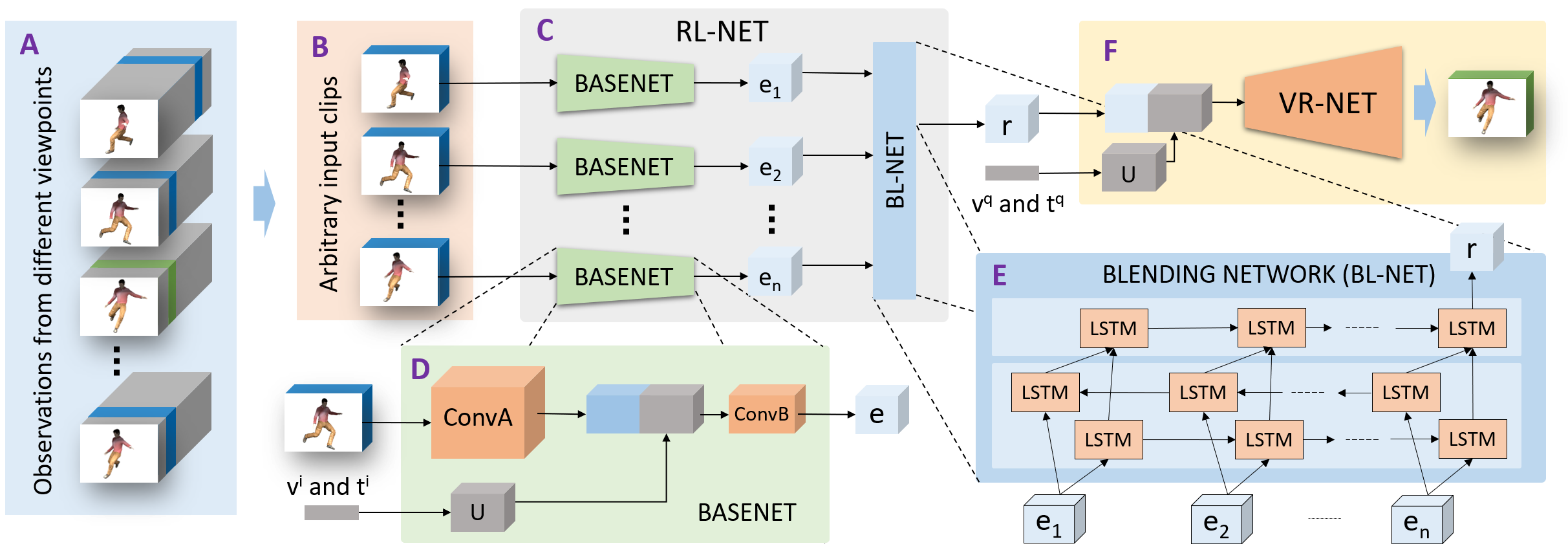}
		\end{center}
		\caption{\small Outline of the proposed view and time-aware video rendering framework. {\bf A} : A collection of observations ($o$) for a given activity from different viewpoints. {\bf B}: Randomly selected clips from the set of observations captured from different viewpoints and at different times. {\bf C}: Representation learning network (RL-NET), which takes video clips from different viewpoint and time as input and learns a representation ($r$). It consists of two parts: Basenet and Blending network. {\bf D}: Basenet is used to learn individual video encodings ($e_i$) conditioned on its viewpoint $v^i$ and time $t^i$ and it is based on 3D convolutions. {\bf E}: The blending network combines encodings learned from different video clips into a unified representation $r$. This blending is performed using a recurrent network which learns a correlation between individual encodings and to get $r$. {\bf F}: Finally, the representation $r$ is used to synthesize a video from arbitrary viewpoint $v^q$ and time $t^q$ using VR-NET which is based on 3D convolution. (ConvA and ConvB refers to convolution and U refers to Upsampling layers.)}
		\label{fig1bf}
	\end{figure*}
	
	\section{Method}
	Following pioneering work of Eslami {\em et. al}\cite {eslami2018neural}, the proposed method consists of two components, a representation learning network, $f$ (RL-NET), and a video rendering network, $g$ (VR-NET). The RL-NET takes multiple video clips captured from varying viewpoints and time of an event termed as observations $o_i = \{(x_i^k, v_i^k, t_i)\}_{k=1,2,...,K}$, where, $x_i^k$ represents $k^{th}$ video clip captured from viewpoint $v_i^k$ and time $t_i$ for any event $i$. The RL-NET take these observations and learns a comprehensive representation, $r_i$, of the event with the help of Blending Network (BL-NET), preserving the view and time notion. BL-NET is a recurrent network maintaining an internal representation, which is updated as more observations are seen by the network before providing a holistic representation $r$ of the scene and its dynamics. The VR-NET, then, use this representation, $r$, along with stochastic latent variables, $z$, to render a video clip from an arbitrary viewpoint, $v_i^k$, and time, $t_i$. 
	
	Formally, we can define the representation learning as, $r_i = f_{\theta}(o_i)$, and the video rendering network as,
	\begin{equation}
		g_{\theta}(x|v^q, t^q, r) = \int g_{\theta}(x, z|v^q, t^q, r)dz,
	\end{equation}
	where, $g_{\theta}(x|v^q, t^q, r)$ represents a probability density of a video $x$ observed from a viewpoint $v^q$ at time $t^q$, for a scene with representation $r$ and latent variable $z$. We train the two networks, RL-NET and VR-NET, jointly in an end-to-end fashion to maximize the likelihood of rendering the ground-truth video, observed from the query viewpoint and timestep. A detailed overview of the proposed framework is shown in Figure \ref{fig1bf}.
	
	\subsection{Representation Learning Network (RL-NET)}
	We use a convolution based neural network to learn the scene representation. RL-NET is shown in Figure  \ref{fig1bf}C, the input consists of multiple video clips along with view-point and time conditioning. The view-point and time conditioning is applied on encodings generated after few convolutions on input clip. We use concatenation of viewpoint, and time with the convolution features from the video for conditioning. It is followed by some more convolution layers. The final encodings from each observation are then combined together to learn a unified scene and dynamics representation, $r$, using a blending network. The encoding network is shared among all the observations.
	
	We explore both 2D and 3D convolution based networks, with conditioning as additional input and generate an encoding for each observation. The learned encodings are, then, used to learn the scene representation, $r$, with size, $(X, Y, Z)$ (Figure \ref{fig1bf}.E). 
	The scene representation is learned using a convolutional recurrent network (BL-NET),  which accumulates information from all the observations effectively (Figure \ref{fig1bf}.E). The work in \cite {eslami2018neural} proposed a simple {\em addition} of encodings to learn the scene representation. We propose a {\em recurrent network}, instead, which we found more effective both in terms of performance and training efficiency. 
	\paragraph{Blending Network (BL-NET)}
	We want to learn a representation which holistically represents the scene, and its dynamics as viewed from varying viewpoints. We propose a recurrent network which updates its representation after looking at each observation (Figure \ref{fig1bf}E). More specifically, we utilize an LSTM architecture \cite{gers1999learning}, where the memory cell, $c$, acts as an accumulator of state information and is updated by the input ($i$), output ($o$) and forget ($f$) gates, which are self-parameterized controlling gates. The order in which the observations are seen by the cell should not have any role in the learned representation, therefore we propose to use bi-directional layers \cite{schuster1997bidirectional} in the network for a more effective learning. Also, to preserve the spatial information in the embeddings, we make use of convolutional LSTM \cite{xingjian2015convolutional}. For a given video embedding, $e_i$, after seeing all other observations in a forward and a backward pass, we get an updated hidden representation $h^r_i$. 
	\begin{equation}
		h^r_i = (o_i^f\circ \tanh(c_i^f))^\frown (o_i^b\circ \tanh(c_i^b)).
	\end{equation}
	Here, $o_i^f$ and $o_i^b$ are the output gates of the forward pass and backward pass respectively, $c_i^f$ and $c_i^b$ are the corresponding memory cell states, $\circ$ denotes the Hadamard product, and $^\frown$ denotes a concatenation operation between learned representations from the forward and backward pass. The updated intermediate representation from each observation is then passed to a uni-directional LSTM layer, which accumulates these to get a holistic representation $r$.
	\begin{equation}
		r = o_n\circ \tanh(c_n).
	\end{equation}
	Here, $o_n$ is the output gate, $c_n$ is the memory cell state of the network after seeing all the $n$ observations.
	\subsection{Video Rendering Network (VR-NET)}
	The representation, $r$, learned based on the given observations, $o$, is used to render a video with a video rendering network (VR-NET). The VR-NET, shown in Figure \ref{fig1bf}F, is also a convolution based network which takes as input the learned representation, $r$, along with query viewpoint, $v^q$, time $t^q$, and latent noise $z$. The viewpoint $v^q$, time $t^q$, and noise $z$ are feed to the network as conditioning, for which we use concatenation operation with the representation features. The VR-NET consists of 2D convolutions followed by 3D convolutions to render the video clips. The convolution layers are used in combination with upsampling of features to generate video clips with resolution same as the input observations. 
	
	The two networks, RL-NET and VR-NET, are trained jointly in an end-to-end fashion minimizing the reconstruction loss $L^r$ as the objective function. The reconstruction loss $L^r$ is computed as mean squared error between the predicted $V^p$ and the ground truth video clip $V^g$.
	\begin{equation}
		L^r = \frac{1}{N}\sum_n^N\sum_i^F \sum_j^H \sum_k^W \sum_m^C ||V^p_{ijkm} - V^g_{ijkm}||^2.
	\end{equation}
	Here, $N$ is the number of samples, $F$ is the number of frames in the clip, $H, W$ is height and width of the video frames, and $C=3$ for three RGB color channels (more details in supplementary file).
	
	\section{Activity Recognition}
	The learned representation can render view-aware as well as time-aware video clips. To further explore the effectiveness of the learned representation, we use it for the task of view-invariant action recognition. We modify the same RL-NET and VR-NET framework and add convolution layers followed by fully connected layers on top of the representation features. This branch of the network (CL-NET) predicts probabilities for each action classes (Figure \ref{fig9}b).  We use categorical cross entropy to compute the loss $L^c$ for the classification branch.
	\begin{equation}
		L^c = -\frac{1}{N}\sum_n^N\sum_c^C 1_{y_i \in C_c} \log(\hat{p}[y_i \in C_c]).
	\end{equation}
	Here, $C$ is the number of action categories, and $\hat{p}[y_i \in C_c]$ is the predicted probability for this sample corresponding to category $c$. The modified network is trained end-to-end with the two loss functions ($L^r$ and $L^c$) in a multi-task setting and the overall loss of the network is defined as,
	\begin{equation}
		L = \lambda1 \times L^r + \lambda2 \times L^c.
	\end{equation}
	In all our experiments, we use $\lambda1=\lambda2=1$. The network is trained using observations captured from certain views and later tested on observations from unseen views. In another variant, we also explore the proposed framework for multi-view action recognition, where all the views are provided during the classification training.
	
	\section{Experiments}
	We perform our experiments on two different real-world datasets: UCF-101 \cite{soomro2012ucf101}, and NTU-RGB+D \cite{shahroudy2016ntu}. UCF-101 does not have the notion of viewpoint, therefore, we use it for time-aware video rendering experiments. NTU-RGB+D is a large scale dataset with videos captured from multiple viewpoints. We use it for view-ware video rendering and view-invariant representation learning, which is further explored for view-invariant activity classification. 
	
	\subsection{Datasets} 
	
	\textbf{UCF-101}:The UCF-101 dataset \cite{soomro2012ucf101} covers a wide range of activities and has around 13K video samples with 101 action classes. There are three different splits in this dataset and we use split-1 for our experiments. \textbf{NTU-RGB+D}: This human  activity  recognition dataset contains more than 56K videos and 4 millions  frames with 60 different actions, including individual activities, interactions between 2 people and health related events. There are a total of 40 different actors, who perform actions captured from 80 different viewpoints. The authors proposed two different splits in this dataset, cross-view and cross-subject. We perform our classification experiments on both the splits as suggested in \cite{shahroudy2016ntu}. Apart from this, we use cross-view split for view-invariant rendering and cross-subject split for other rendering experiments.
	
	\subsection{Training}
	We perform the pre-processing of video frames in UCF-101 as suggested in \cite{tran2015learning}, and take a random crop of 112x112 on the resized frames. In the time-aware frame rendering experiments, the RL-NET takes 6 frames in a video randomly  and the VR-NET generates a video frame from arbitrary time in the clip. In video rendering experiments, the input video clips have 6 frames and we feed in 3 randomly selected clips to the RL-NET for representation learning, and VR-NET generates a video clip with 6 frames from arbitrary time.
	
	In all our experiments on NTU-RGB+D dataset, we use subject split, except for the view-split based classification task. For view-based rendering, the subject training split is further divided to get the desired view. We resize the video frames to 240x135, preserving the aspect ratio, and then take a random 112x112 crop for training. The input clips with 6 frames are randomly selected from each view, with varying time, and are used for representation learning, and subsequently the VR-NET generates a video clip with 6 frames from arbitrary view and time. The additional view-point information is used along with time for representation learning as well as video rendering.
	
	In all our experiments we use Adam optimizer \cite{kingma2014adam} with a learning rate of 1e-4 and a batch size of 6. We implemented our code in Keras with Tensorflow backend and use Titan-X GPU for training our network.

	\begin{figure}[t!] 
		\includegraphics[width=0.15\linewidth]{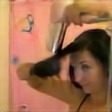}
		\includegraphics[width=0.15\linewidth]{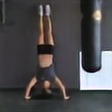}
		\includegraphics[width=0.15\linewidth]{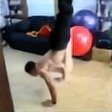}
		\includegraphics[width=0.15\linewidth]{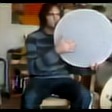}
		\includegraphics[width=0.15\linewidth]{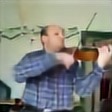}
		\includegraphics[width=0.15\linewidth]{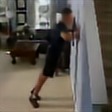}  \\
		\includegraphics[width=0.15\linewidth]{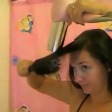}
		\includegraphics[width=0.15\linewidth]{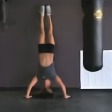}
		\includegraphics[width=0.15\linewidth]{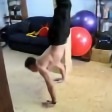}
		\includegraphics[width=0.15\linewidth]{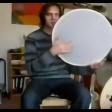}
		\includegraphics[width=0.15\linewidth]{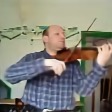}
		\includegraphics[width=0.15\linewidth]{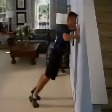}  \\
		\includegraphics[width=0.15\linewidth]{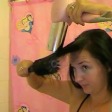}
		\includegraphics[width=0.15\linewidth]{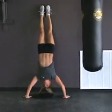}
		\includegraphics[width=0.15\linewidth]{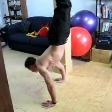}
		\includegraphics[width=0.15\linewidth]{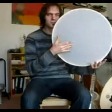}
		\includegraphics[width=0.15\linewidth]{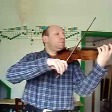}
		\includegraphics[width=0.15\linewidth]{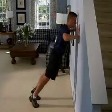}
		\caption{\small Images generated with time conditioning using 2D-CNN representation on UCF-101. Row-1: LSTM generator and, Row-2: CNN generator and Row-3: Ground truth.}
		\label{fig3} 
	\end{figure}
	
	\subsection{Time-aware Rendering}
	\label{l_tar}
	Building upon view-invariant neural representation for image generation using synthetic data \cite{eslami2018neural}, we learn a representation encompassing the temporal domain. The video representation $r$ is learned conditioned on the time factor using a convolution network (RL-NET). The representation $r$, with an understanding of the temporal domain, is used to generate an image/clip at arbitrary time in the video with the help on VR-NET.
	
	In their work \cite{eslami2018neural}, related to view-invariant {\em image} representation, the authors use an iterative LSTM generator. Drawing on similar lines, we explore a recurrent LSTM generator (VR-NET) for image generation. Apart from this, we also explore a CNN generator (VR-NET) with similar training conditions. The images generated by both the networks are shown in Figure \ref{fig3} for a comparison. In the absence of any marked difference between the quality of generated images, we choose a CNN generator (VR-NET) for further experiments, which is faster to train in comparison with an LSTM network. The representation learned by this network can also predict 3-4 frames in future or the whole video frame by frame. A comparison of the generated image quality with other methods is shown in Table \ref{table4}. We use the same testing setup as suggested in \cite{byeon2018contextvp}. The method proposed by \cite{byeon2018contextvp} performs better in terms of PSNR values, however, it is important to note that this method was specifically trained for future frame prediction, whereas our network was trained for predicting arbitrary frames in the video. Moreover, our network outperforms all the other methods in terms of SSIM score.
	
	\begin{table}[t!]
		\small
		\begin{center}
			\begin{tabular}{|l|p{15mm}|p{15mm}|p{20mm}|}
				\hline
				Method & PSNR & SSIM \\
				\hline\hline
				\cite{mathieu2015deep} & 32 & 0.92  \\
				\hline
				\cite{villegas2017decomposing} & 31 & 0.91 \\
				\hline
				\cite{liu2017video} & 33.4 & 0.94 \\
				\hline
				\cite{byeon2018contextvp} & \textbf{34.9} & 0.92 \\
				\hline
				Ours & 31.9 & \textbf{0.95} \\
				\hline
			\end{tabular}
		\end{center}
		\caption{\small Comparison of the quality of future-frame prediction on 10\% of UCF-101 test dataset. The network is provided 4 frames to predict the next frame. It is important to note that the other methods were trained specifically to predict the future frame whereas our network was trained to predict any arbitrary frame in the video.}
		\label{table4}
	\end{table}
	
\begin{figure}[t!]
		\centering
		\includegraphics[width=0.13\linewidth]{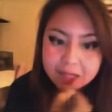}
		\includegraphics[width=0.13\linewidth]{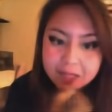}
		\includegraphics[width=0.13\linewidth]{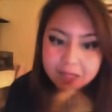}
		\includegraphics[width=0.13\linewidth]{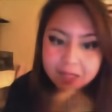}
		\includegraphics[width=0.13\linewidth]{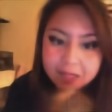}
		\includegraphics[width=0.13\linewidth]{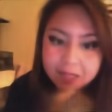}\\
		\includegraphics[width=0.13\linewidth]{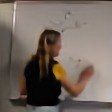}
		\includegraphics[width=0.13\linewidth]{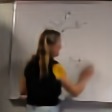}
		\includegraphics[width=0.13\linewidth]{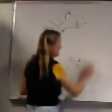}
		\includegraphics[width=0.13\linewidth]{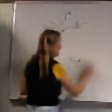}
		\includegraphics[width=0.13\linewidth]{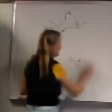}
		\includegraphics[width=0.13\linewidth]{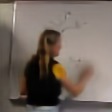}\\
		\includegraphics[width=0.13\linewidth]{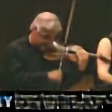}
		\includegraphics[width=0.13\linewidth]{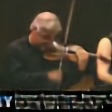}
		\includegraphics[width=0.13\linewidth]{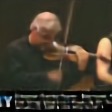}
		\includegraphics[width=0.13\linewidth]{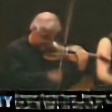}
		\includegraphics[width=0.13\linewidth]{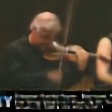}
		\includegraphics[width=0.13\linewidth]{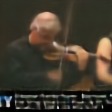}\\
		\includegraphics[width=0.13\linewidth]{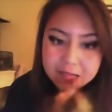}
		\includegraphics[width=0.13\linewidth]{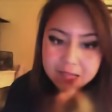}
		\includegraphics[width=0.13\linewidth]{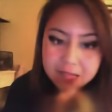}
		\includegraphics[width=0.13\linewidth]{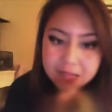}
		\includegraphics[width=0.13\linewidth]{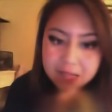}
		\includegraphics[width=0.13\linewidth]{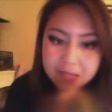}\\
		\includegraphics[width=0.13\linewidth]{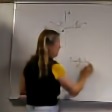}
		\includegraphics[width=0.13\linewidth]{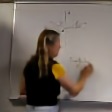}
		\includegraphics[width=0.13\linewidth]{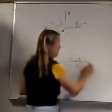}
		\includegraphics[width=0.13\linewidth]{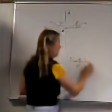}
		\includegraphics[width=0.13\linewidth]{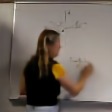}
		\includegraphics[width=0.13\linewidth]{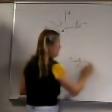}\\
		\includegraphics[width=0.13\linewidth]{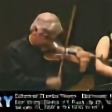}
		\includegraphics[width=0.13\linewidth]{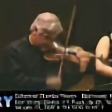}
		\includegraphics[width=0.13\linewidth]{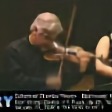}
		\includegraphics[width=0.13\linewidth]{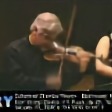}
		\includegraphics[width=0.13\linewidth]{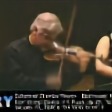}
		\includegraphics[width=0.13\linewidth]{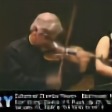}
		\caption{\small Future video frames rendered for UCF101 dataset, with a time conditioned 3D convolution VR-NET; top-three-rows are generated with: 2D convolution RL-NET, Row-1: Apply lipstick, Row-2: Writing on board, and Row-3: Playing violin; and bottom-three-rows use 3D convolution RL-NET: Row-4: Apply lipstick, Row-5: Writing on board, and Row-6: Playing violin.}
		\label{fig4}
	\end{figure}

	Next, we explore the video generation capability of the proposed framework. We experimented with both 2D and 3D convolutions to learn a representation with 3D-CNN based RL-NET. The input to RL-NET is a set of frames from arbitrary time in case of 2D-CNN and a set of video clips in case of 3D-CNN. The frames from the generated videos for some of the UCF-101 action classes are shown in Figure \ref{fig4}. The videos generated using 3D-CNN based RL-NET are relatively better when compared with 2D-CNN. We compared our method with \cite{mathieu2015deep} for future video prediction and observed an improvement in the predicted video quality based on PSNR and SSIM measures (Table \ref{table3}).
	
	\begin{table}
		\small
		\begin{center}
			\begin{tabular}{|c|c|c|c|c|c|c|}
				\hline
				\multirow{2}{*}{} & \multicolumn{2}{c|}{FF} & \multicolumn{2}{c|}{LF} & \multicolumn{2}{c|}{FV} \\
				\cline{2-7}
				& PSNR & SSIM & PSNR & SSIM & PSNR & SSIM \\
				\hline \hline
				\cite{mathieu2015deep} & 21.4 & 0.69 & 17.7 & 0.58 & - & - \\
				\hline
				Ours & \textbf{24.68} & \textbf{0.79} & \textbf{21.66} & \textbf{0.67} & \textbf{22.97} & \textbf{0.73} \\
				\hline \hline
				O-1 & 19.9 & 0.68 & 19.8 & 0.68 & 19.8 & 0.68 \\
				\hline
				O-2 & 19.9 & 0.69 & 19.9 & 0.69 & 19.8 & 0.68 \\
				\hline
				O-3 & 23.3 & 0.79 & 23.2 & 0.79 & 23.3 & 0.79 \\
				\hline
			\end{tabular}
		\end{center}
		\caption{\small Quantitative evaluation of the future video predictions on 10\% UCF-101 and time as well as view-aware prediction on NTU-RGB+D test set. The first two rows are for UCF-101 and the bottom 3 rows are our results on NTU-RGB+D dataset. FF: first frame, LF: last frame, and FV: full video. O-1: input is 2-random views from (1,2,3) and output is the left out view, O-2: input is views 2 and 3 and output is view 1, O-3: input is from all 3 views and output is from random view. The query time is randomly chosen for all the experiments.}
		\label{table3}
	\end{table}

	\subsection{View Invariant Video Generation}
	\label{l_var}
	
	We perform our view-invariant video rendering experiments on NTU-RGB+D dataset. We train the network with view-point conditioning along with time condition to render videos from arbitrary view-points. A viewpoint is defined using 6 different parameters: camera height, camera distance, view-point, horizontal-pan, vertical-pan, and actor position (refer to supplementary files for more details).

	\begin{figure}[t!]
		\centering
		\includegraphics[width=0.15\linewidth]{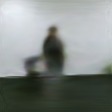}
		\includegraphics[width=0.15\linewidth]{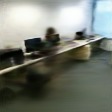}
		\includegraphics[width=0.15\linewidth]{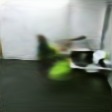}
		\includegraphics[width=0.15\linewidth]{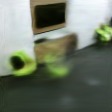}
		\includegraphics[width=0.15\linewidth]{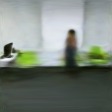}
		\includegraphics[width=0.15\linewidth]{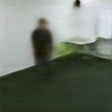}
		\\
		\includegraphics[width=0.15\linewidth]{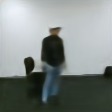}
		\includegraphics[width=0.15\linewidth]{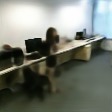}
		\includegraphics[width=0.15\linewidth]{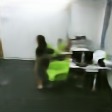}
		\includegraphics[width=0.15\linewidth]{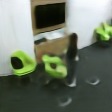}
		\includegraphics[width=0.15\linewidth]{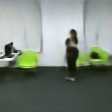}
		\includegraphics[width=0.15\linewidth]{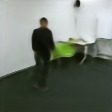}
		\\
		\includegraphics[width=0.15\linewidth]{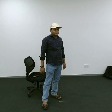}
		\includegraphics[width=0.15\linewidth]{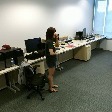}
		\includegraphics[width=0.15\linewidth]{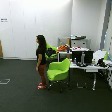}
		\includegraphics[width=0.15\linewidth]{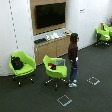}
		\includegraphics[width=0.15\linewidth]{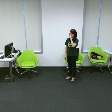}
		\includegraphics[width=0.15\linewidth]{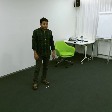}
		\caption{\small Images generated with time and view conditioning using 2D-CNN network on NTU-RGB+D dataset, where representation uses a 2D-CNN network with 2 different blending techniques: Row-1: Addition; Row-2: BL-NET; and Row-3: Ground truth.}
		\label{fig6}
	\end{figure}
	
	\begin{figure}[t!]
		\begin{center}
			\includegraphics[width=0.48\linewidth]{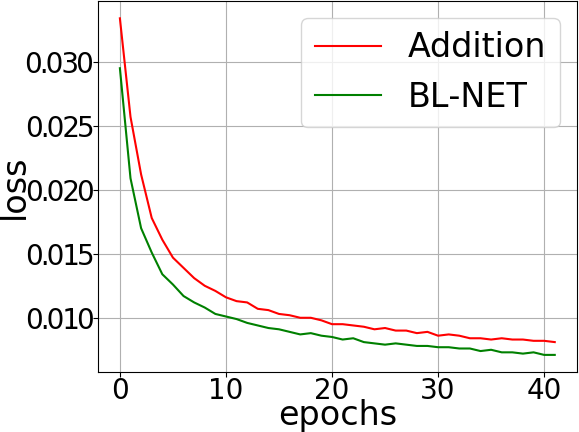} ~~
			\includegraphics[width=0.48\linewidth]{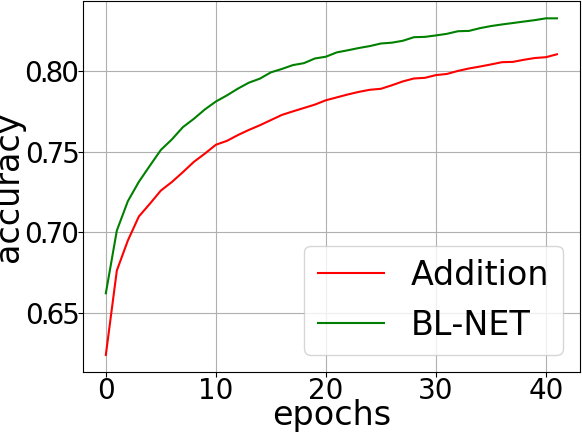}
		\end{center}
		\caption{\small A performance comparison between two blending techniques: Addition and BL-NET. Left plot shows the change in loss and the right plot shows the change in accuracy as training progresses. The accuracy is computed using per-pixel prediction at a certain threshold. We observe a faster convergence and a better performance with BL-NET as compared to addition blending.}
		\label{fig_blending}
	\end{figure}
	
	We first explore the effectiveness of the proposed blending network (BL-NET) for representation learning. We use a 2D-CNN based RL-NET along with proposed BL-NET to learn a representation $r$ for a given set of frames captured from different view-points and time. The learned representation $r$ is used then used to render a frame from arbitrary viewpoint and time. We perform similar experiment with addition in place of BL-NET for representation learning. Some of the generated images with these experiments are shown in Figure \ref{fig6}. We can observe that the quality of generated images is much better with BL-NET representation. Apart from this, we also compare the variation in loss and learning curve for these two blending techniques (Figure \ref{fig_blending}). We observe a faster learning and a better performance for BL-NET.
	
	
	
	\begin{figure}[t!]
		\begin{center}
			\includegraphics[width=0.95\linewidth]{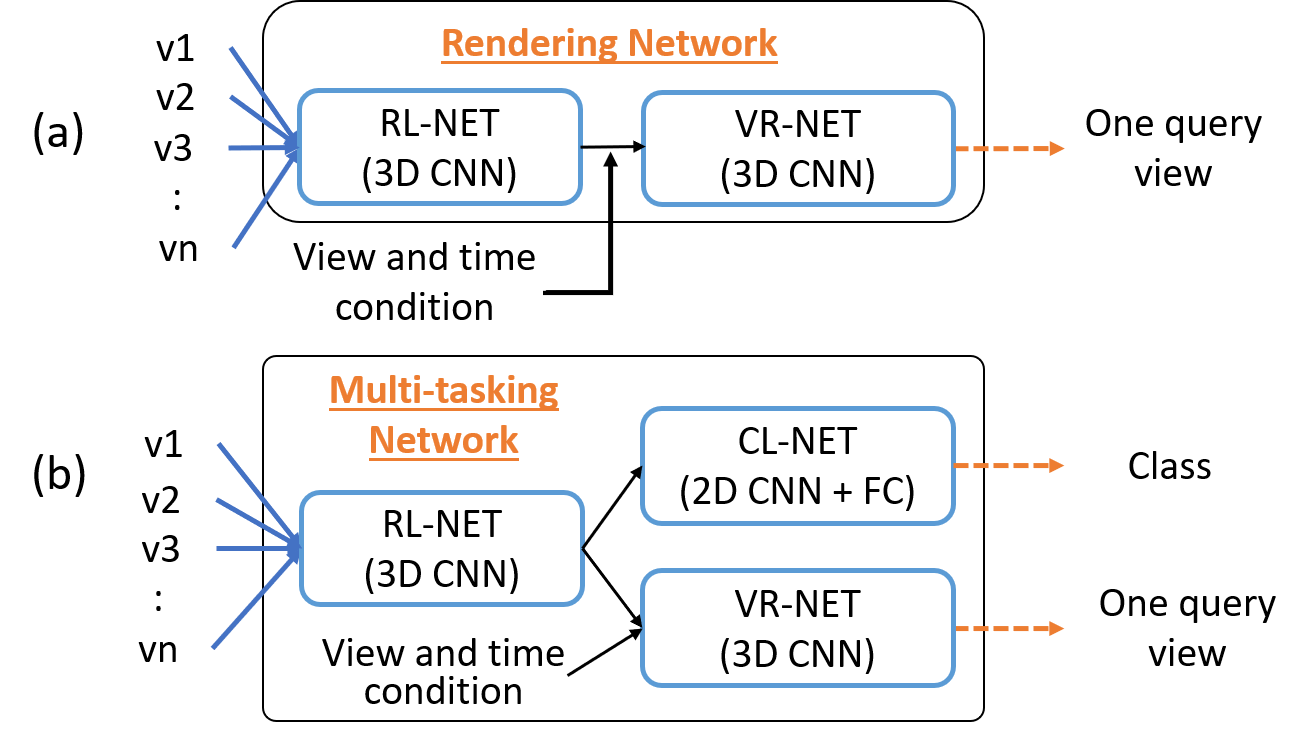}
		\end{center}
		\caption{\small (a) Generalized network architecture for video rendering using 3D-CNN RL-NET and VR-NET. (b) Multitasking network for classification along with video rendering. 
		}
		\label{fig9}
	\end{figure}
	
\begin{figure}[t!]
		\centering 
		\includegraphics[width=0.15\linewidth]{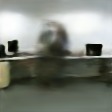}
		\includegraphics[width=0.15\linewidth]{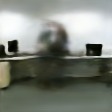}
		\includegraphics[width=0.15\linewidth]{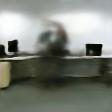}
		\includegraphics[width=0.15\linewidth]{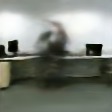}
		\includegraphics[width=0.15\linewidth]{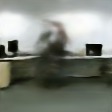}
		\includegraphics[width=0.15\linewidth]{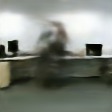}\\
		
		\includegraphics[width=0.15\linewidth]{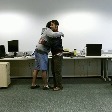}
		\includegraphics[width=0.15\linewidth]{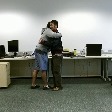}
		\includegraphics[width=0.15\linewidth]{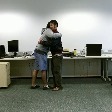}
		\includegraphics[width=0.15\linewidth]{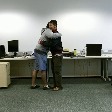}
		\includegraphics[width=0.15\linewidth]{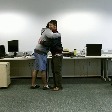}
		\includegraphics[width=0.15\linewidth]{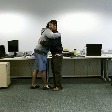}\\
		
		\includegraphics[width=0.15\linewidth]{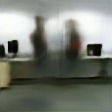}
		\includegraphics[width=0.15\linewidth]{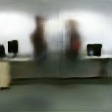}
		\includegraphics[width=0.15\linewidth]{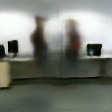}
		\includegraphics[width=0.15\linewidth]{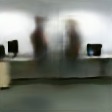}
		\includegraphics[width=0.15\linewidth]{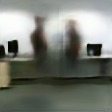}
		\includegraphics[width=0.15\linewidth]{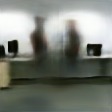}\\
		
		\includegraphics[width=0.15\linewidth]{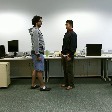}
		\includegraphics[width=0.15\linewidth]{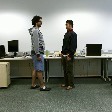}
		\includegraphics[width=0.15\linewidth]{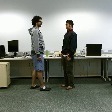}
		\includegraphics[width=0.15\linewidth]{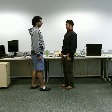}
		\includegraphics[width=0.15\linewidth]{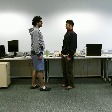}
		\includegraphics[width=0.15\linewidth]{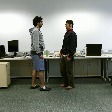}\\
		
		\includegraphics[width=0.15\linewidth]{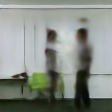}
		\includegraphics[width=0.15\linewidth]{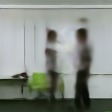}
		\includegraphics[width=0.15\linewidth]{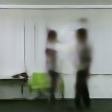}
		\includegraphics[width=0.15\linewidth]{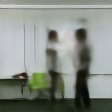}
		\includegraphics[width=0.15\linewidth]{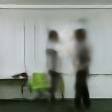}
		\includegraphics[width=0.15\linewidth]{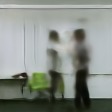}\\
		
		\includegraphics[width=0.15\linewidth]{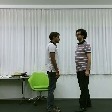}
		\includegraphics[width=0.15\linewidth]{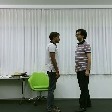}
		\includegraphics[width=0.15\linewidth]{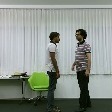}
		\includegraphics[width=0.15\linewidth]{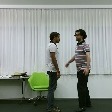}
		\includegraphics[width=0.15\linewidth]{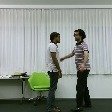}
		\includegraphics[width=0.15\linewidth]{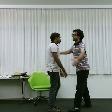}
		\caption{\small Generated and ground truth video frames (action class- 55, hugging other person) at different time-steps generated using models trained under different conditions. Row-1-2: Given two random views generates third unseen view, Row-3-4: Given view 2 and 3 generates unseen view 1, and Row-5-6: Given all three views generates one given view at an arbitrary query time.}
		\label{renderingNTU}
	\end{figure}

\begin{figure}[t!]
		\centering
		\includegraphics[width=0.15\linewidth]{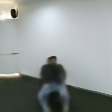}
		\includegraphics[width=0.15\linewidth]{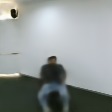}
		\includegraphics[width=0.15\linewidth]{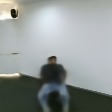}
		\includegraphics[width=0.15\linewidth]{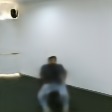}
		\includegraphics[width=0.15\linewidth]{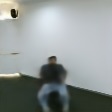}
		\includegraphics[width=0.15\linewidth]{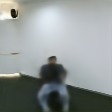}\\
		
		\includegraphics[width=0.15\linewidth]{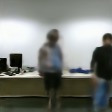}
		\includegraphics[width=0.15\linewidth]{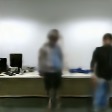}
		\includegraphics[width=0.15\linewidth]{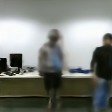}
		\includegraphics[width=0.15\linewidth]{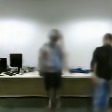}
		\includegraphics[width=0.15\linewidth]{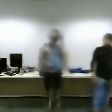}
		\includegraphics[width=0.15\linewidth]{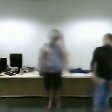}\\
		
		\includegraphics[width=0.15\linewidth]{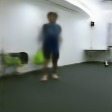}
		\includegraphics[width=0.15\linewidth]{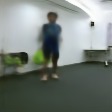}
		\includegraphics[width=0.15\linewidth]{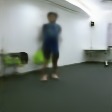}
		\includegraphics[width=0.15\linewidth]{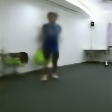}
		\includegraphics[width=0.15\linewidth]{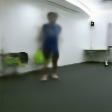}
		\includegraphics[width=0.15\linewidth]{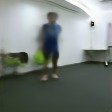}
		\caption{\small Video frames generated with time and view conditioned network with three views. Video generated contains future frames for one of the given views and action classes, Row-1: class 18(wear on glasses) , Row-2: class 57(touch other person's pocket), and Row-3: class 43(falling).}
		\label{fig10}
	\end{figure}

	
	We use a 3D-CNN based RL-NET along with BL-NET blending to learn $r$ for a given set of video clips from different viewpoints and time, which is then passed to a 3D-CNN based VR-NET for video rendering from an arbitrary viewpoint and time. (Figure \ref{fig9}a). We experimented with three different variations in the training setup based on the input and rendered videos (Figure \ref{renderingNTU}). We start with a challenging setup, where the input clips are from $3 \choose 2$ (1,2,3), i.e. two randomly selected views from three views (1, 2, 3) and VR-NET renders the third unseen view (Figure \ref{renderingNTU} row 1 \& 2). In the second variation, the input clips are from view 2 and 3 and VR-NET always generates view 1 (row 3 \& 4). In the third variation, the input was given from all three views and VR-NET renders a video from a randomly selected view among the three views (row 5 \& 6). We observe that all the three variations generate video-clips with correct view-point on the test-set. Although we observe some blur in the regions with motion, the quality of the videos generated is best with the third variation. Some more synthesized  video frames for this variation are shown in Figure \ref{fig10}.
	
	We also perform a quantitative evaluation of the proposed view and time aware rendering approach. We compute PSNR and SSIM measures to evaluate the quality of generated videos. The evaluation scores are shown in row 3-5 of Table \ref{table3}. We observe that the quality of generated videos is better when all the views are seen by the network as compared with generating unseen views which was expected.
	
	\subsection{Activity Classification}
	\label{l_ar}
	We further explore the effectiveness of the proposed representation learning approach for view-invariant activity classification. We modify the proposed framework for multi-tasking (Figure \ref{fig9}b) where the network is jointly trained for video rendering as well as activity classification. Although, multi-tasking with classification does not improve rendering quality, we obtain state-of-the-art activity classification results on NTU-RGB+D dataset using RGB modality. We experimented with different training variations based on the input to RL-NET (different combinations of views and number of clips) and query to VR-NET.
	
\begin{figure}[t!]
		\begin{center}
			\includegraphics[width=0.98\linewidth]{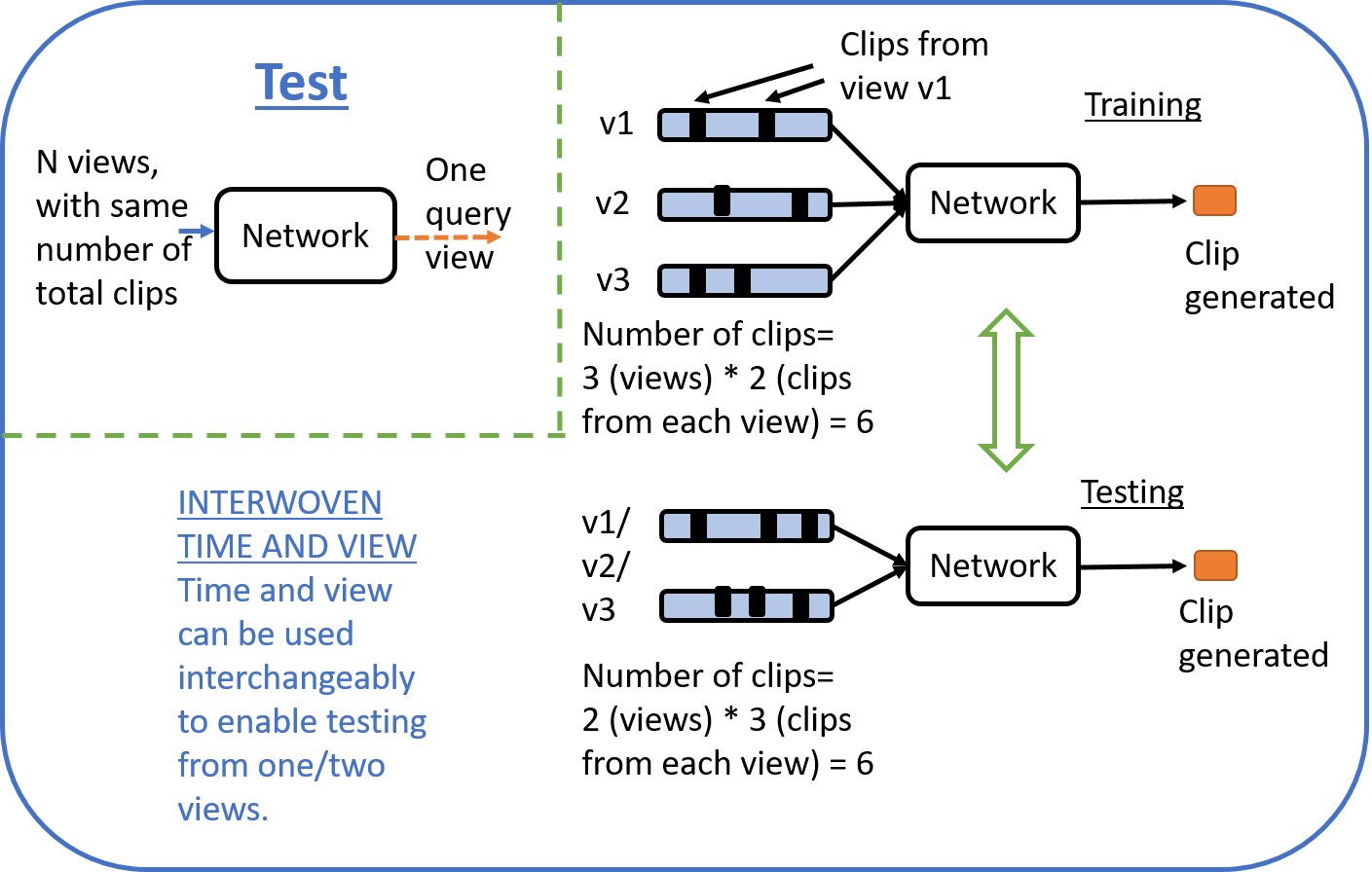}
		\end{center}
		\caption{Interwoven time and view dimensions: time and view dimensions can be used interchangeably while making predictions with our network.}
		\label{fig11}
	\end{figure}

	The RL-NET performs representation learning based on a set of input observations. These observations can be from different viewpoints (varying views) and different time in a video (Figure \ref{fig11}). The visual appearance of any activity changes with viewpoint as well as time. This analogy between time and viewpoint for variation in visual appearance of any activity allows us to use the two concepts interchangeably during representation learning. This idea makes the proposed architecture even more powerful as a network trained under some settings can be tested on different set of parameters in terms of number of views and number of clips. The proposed BL-NET supports this further due to its recurrent structure which allows it to learn a representation for different varying observations.
	
	We perform an ablation study on activity classification to explore this idea further. In Table \ref{table1}, we show the classification accuracy on a cross subject-split using three different network variations. The classification accuracy is highest, when all three views are available which provides more activity information. Please note that the actors are either facing view 2 or view 3 in this dataset, which provides uniformity to the input (2, 3) thus accuracy is second highest when we use this input view. 
	
	\begin{table}[t!]
		\small
		\begin{center}
			\begin{tabular}{|c|p{10mm}|p{10mm}|p{15mm}|}
				\hline
				\multirow{2}{*}{Testing View} & \multicolumn{3}{c|}{Training View} \\
				\cline{2-4}
				& (1,2,3) & (2,3) & $3 \choose 2$ (1,2,3)\\
				\hline\hline
				(1,2,3) & \textbf{67.36}  & \textbf{78.06} & \textbf{83.79} \\
				\hline
				(1,2) & 62.88 & 62.0 & 80.44\\
				\hline
				(2,3) & 65.96 & 69.6 & 80.67 \\
				\hline
				(1,3) & 62.01 & 63.1 & 80.09\\
				\hline
				$3 \choose 2$ (1,2,3) & 63.13 & 64.93 & 79.67 \\
				\hline
				(1) & 57.28 & 59.8 & 77.26 \\
				\hline
				(2) & 56.16 & 59.7 & 74.17 \\
				\hline
				(3) & 55.25 & 58.4 & 72.23 \\
				\hline
			\end{tabular}
		\end{center}
		\caption{\small Ablation experiments to study the effectiveness of the learned representation. The table shows classification accuracy with different trained networks and inputs. Variation in the training and testing views and number of input clips leads to different classification accuracy. Here, $3 \choose 2$ (1,2,3) refers to two randomly picked views from the three views (1,2,3).}
		\label{table1}
	\end{table}
	
\begin{figure}[t!]
		\begin{center}
			\includegraphics[width=0.75\linewidth]{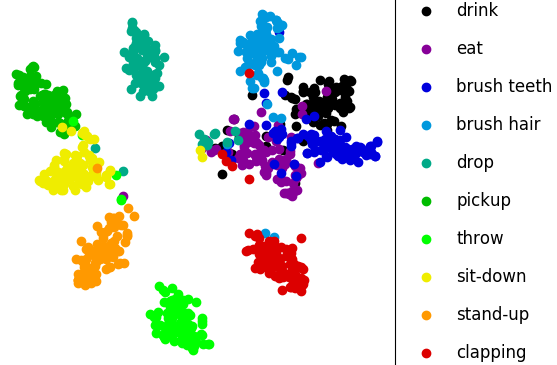}
		\end{center}
		\caption{t-SNE visualization of activity representations for a subset of 10 activities (out of 60) on NTU-RGB+D dataset. The representation is learned with two random input views to RL-NET where VR-NET generates the third view and activity classification is performed. Most of the actions are well separated and the actions with similar appearance and dynamics are close to each other.}
		\label{fig12}
	\end{figure}

	We also observe that the representation learned with two random input views ($3 \choose 2$ (1,2,3)) provides the highest classification accuracy of 83.79\%. {\em When network generates the third unseen view during the training the representation learned is more robust}. The t-SNE embedding for this model are shown in the Figure \ref{fig12}. We observe that the learned embedding for a given class are clustered together and the activity classes which are visually similar are close to each other.
	
	We compare the performance of our proposed method with the recent works on view-invariant activity recognition (Table \ref{table2}). We observe that the proposed method outperforms the existing methods with a significant margin ($\sim$ 26\%). Moreover, the performance is comparable with the state-of-the-art approaches employing depth and skeleton modalities \cite{luo2017unsupervised}\cite{li2018unsupervised}. Our model performs well in both cross-subject (79.23\% with one view) and cross-view split (75.59\% with unseen view 1) and outperforms the state-of-the-art methods using only RGB modality. The weight sharing form of our network provides great flexibility in increasing the input clips to our network. Thus, when we increase the number of input clips from a video, the best subject-split classification accuracy of 83.79\% with three views (Table \ref{table1}) increases to 86.78\%, when we input multiple clips.


	\begin{table}[t!]
		\small
		\centering
		\begin{center}
			\begin{tabular}{|l|p{15mm}|p{10mm}|p{10mm}|}
				\hline
				Method & Modality & Cross-subject & Cross-view \\
				\hline\hline
				Luo et al. \cite{luo2017unsupervised} & D & 66.2 & 53.2 \\
				\hline
				Li et al. \cite{li2018unsupervised} & D & 68.1 & 63.9 \\
				\hline
				Shuffle \& learn \cite{10.1007/978-3-319-46448-0_32} & D & 61.4 & 53.2 \\
				\hline \hline
				Shahroudy et al.\cite{shahroudy2016ntu} & S & 62.9 & 70.3 \\
				\hline \hline
				DSSCA - SSLM\cite{shahroudy2018deep} & RGB-D-S  & 74.9 & - \\
				\hline \hline
				Luo et al. \cite{luo2017unsupervised} & RGB & 56 & - \\
				\hline
				Li et al. \cite{li2018unsupervised} & RGB & 55.5 & 49.3 \\
				\hline
				Proposed & RGB & \textbf{79.23} & \textbf{75.59} \\
				\hline
				Proposed-AV & RGB & \textbf{86.78} & \textbf{-} \\
				\hline
			\end{tabular}
		\end{center}
		\caption{\small Quantitative evaluation in terms of classification accuracy on NTU-RGB+D dataset. Note that rows from 1-5 use different modalities such as depth (D) and/or skeleton (S) in addition to RGB; we only use RGB. Proposed-AV method takes all the views at once for classification and we observe a significant improvement with this prior on testing samples.
		}
		\label{table2}
	\end{table}

	
	\section{Conclusion and Future Work}
	In this work, we introduce a novel view and time aware video rendering problem. We propose a simple deep learning based framework to solve this problem. The proposed framework consists of two components, representation learning network (RL-NET): which learns a robust representation of a time-varying event captured from multiple viewpoints, and a video rendering network (VR-NET) which synthesizes a video for a given viewpoint and time of the same event. We demonstrate the effectiveness of the proposed framework to render view and time aware videos on two different real world datasets. The video rendering network forces the representation network to learn a more robust and effective representation of the event. To validate this further, we utilize the learned representation for view-aware activity classification task and show state-of-the-art results on NTU-RGB+D dataset. We believe the idea of rendering unseen view is really powerful and it enables the network to learn a meaningful representation which can be used for multiple other tasks. In this work, we mainly focused on visual data and we plan to explore this further for other modalities and problem domains.    
	
	\subsection*{Acknowledgments}
	This research is based upon work supported in parts by the National Science Foundation under Grant No. 1741431; and the Office of the Director of National Intelligence (ODNI), Intelligence Advanced Research Projects Activity (IARPA), via IARPA R\&D Contract No. D17PC00345. The views, findings, opinions, and conclusions or recommendations contained herein are those of the authors and should not be interpreted as necessarily representing the official policies or endorsements, either expressed or implied, of the NSF, ODNI, IARPA, or the U.S. Government. The U.S. Government is authorized to reproduce and distribute reprints for Governmental purposes notwithstanding any copyright annotation thereon.
	
	{\small
		\bibliographystyle{ieee}
		\bibliography{egbib}
	}
	
	\newpage
	
	\appendix
	\section{Appendices}
	
	This document provides additional details supplementary to the main manuscript. It covers network architectures, training details and some additional results which were not included in the main manuscript.
	
	\section{Model Details}
\begin{figure}[htp]
		\begin{center}
			\includegraphics[width=0.5\linewidth]{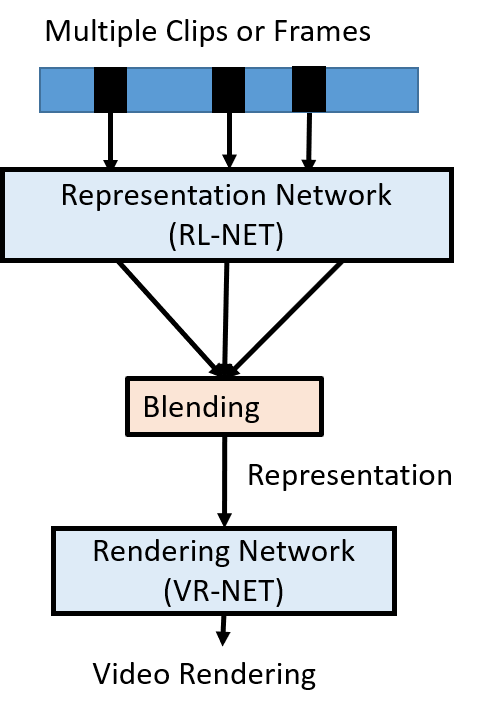}
		\end{center}
		\caption{An outline of the proposed framework for representation learning and video rendering for single view videos.}
		\label{out}
	\end{figure}

	The proposed approach consists of two main components: Representation Learning network (RL-NET) and Video Rendering Network (VR-NET) (shown in  Figure \ref{out}). RL-NET is used to learn a representation $r$ given some observations $o_i$. It consists of a base network (BASENET), which is shared among all the observations to encode individual observations, and a Blending Network (BL-NET), which aggregates encodings from all the observations to learn a unified representation $r$. The representation $r$ is then passed to the VR-NET which renders a video from arbitrary view and time. The multi-tasking architecture, which also performs action classification, consists of another branch caled Classification Network (CL-NET). It takes the learned representation $r$ as input and predict confidence scores for the activity classes. We will discuss the architecture details for all these networks in the following subsections.
	
	\subsection{Representation Learning Network (RL-NET)}
\begin{figure*}[ht!]
		\begin{center}
			\includegraphics[width=0.95\linewidth]{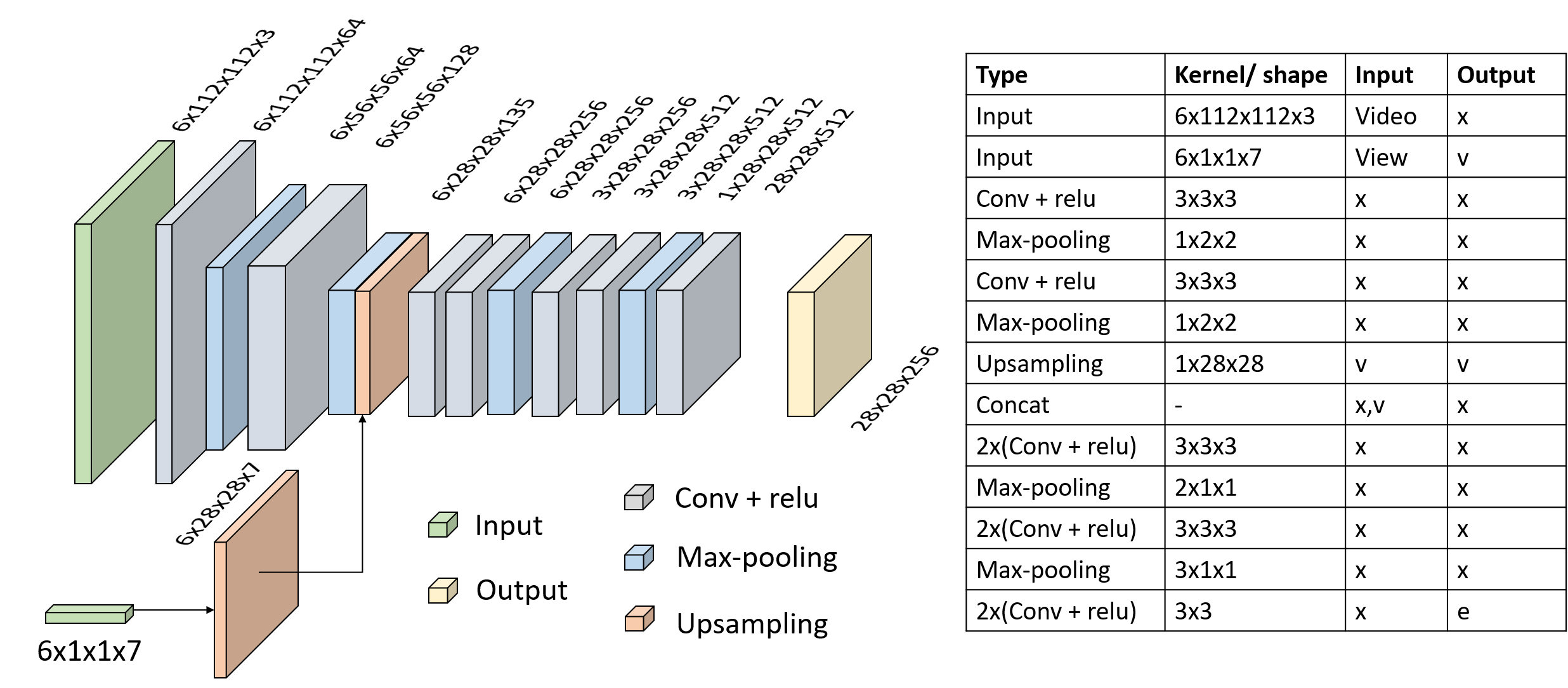}
		\end{center}
		\caption{Network architecture of BASENET used in RL-NET for representation learning $r$ for set of video observations $o_i$. The diagram on the left side shows the change in feature volumes, along with size, as the input to the network flows through the layers of the network. The network takes two input, video clips and corresponding viewpoints, and generates a video. The table on the right shows details of the layers in the network including kernel sizes and layer types. Conv: 3D convolution in all the layers except the last two layers where it will be 2D convolution as the temporal extent has been compressed to single channel, v: viewpoint, e: embeddings for observations.}
		\label{basenet_3d}
	\end{figure*}

	The overview of the RL-NET is shown in Figure 2.C (main manuscript). BASENET is the encoding part of RL-NET which is shared among all the observations and the corresponding encodings are passed to the BL-NET for representation learning. We experimented with two different networks for representation learning. We first explored 2D convolutions for single view videos (UCF-101), where each observation is a single video frame. The BASENET for this architecture utilizes 2D convolution in combination with max-pooling to get frame level encodings. We further extend this to short video clips where each observation is a set of consecutive video frames with some skip rate. In this case we use 3D convolutions instead of 2D and the structure of the architecture remains almost similar. A detailed architecture for this variation is shown in Figure \ref{basenet_3d}. The 2D variant of representation learning network has similar architecture with 3D convolutions substituted with 2D convolutions.
	
	The encodings learn by the base network are passed on to the blending network (BL-NET), which is a recurrent network with bi-directional convolutional LSTM cells. The architecture of BL-NET is shown in Figure 2.E (main manuscript). The number of recurrent steps in the network will be equal to the number of observations ($o_i$) provided to the RL-NET. The first layer of BL-NET comprise of 128 3x3 kernels in each direction with a total of 256 kernels for the convolution operation. The final layer has 256 3x3 kernels which produces a representation of size 28x28x256.
	
	\begin{figure*}[ht!]
		\begin{center}
			\includegraphics[width=0.95\linewidth]{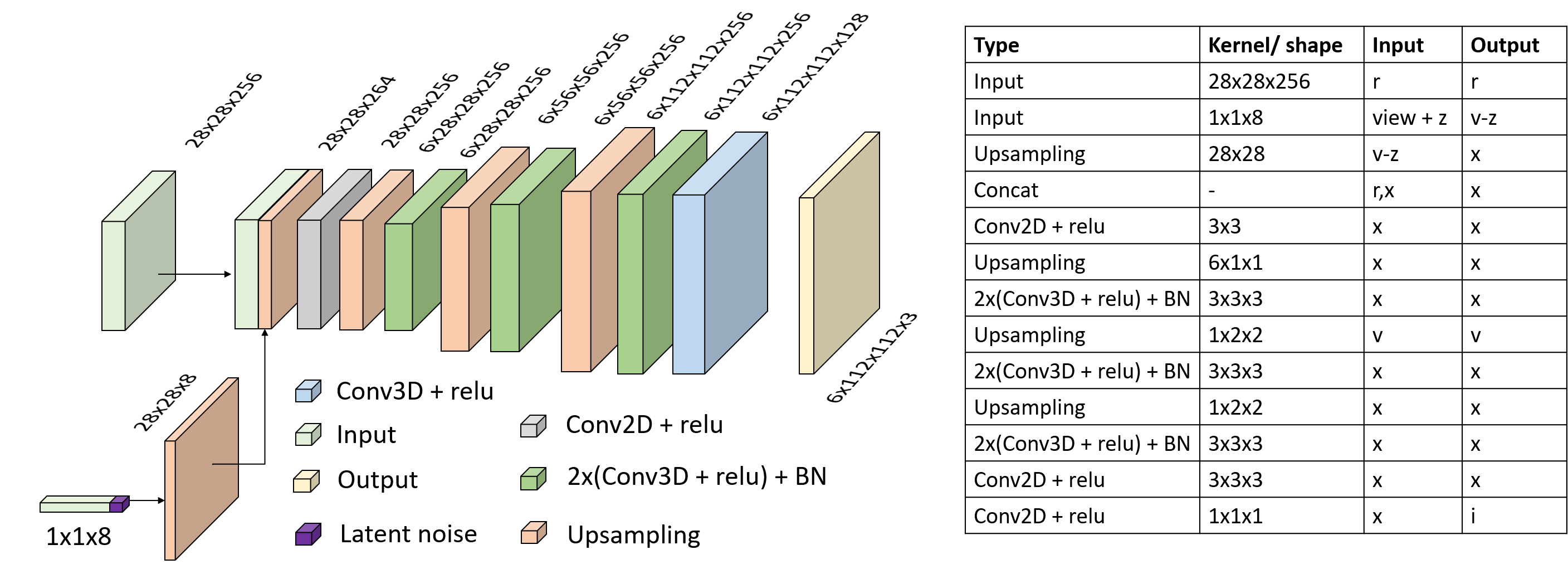}
		\end{center}
		\caption{Network architecture of VR-NET used for video rendering. VR-NET takes the learned representation $r$ along with query viewpoint $v$ and latent noise $z$ as input and generates a video. The representation $r$, viewpoint $v$ and latent noise $z$ are first integrated together followed by a convolution operation. A video is generated using 3D convolution along with batch-normalization and upsampling. BN: batch normalization, z: latent noise, i: generated video.}
		\label{vrnet_3d}
	\end{figure*}
	
	\begin{figure*}[ht!]
		\begin{center}
			\includegraphics[width=0.7\linewidth]{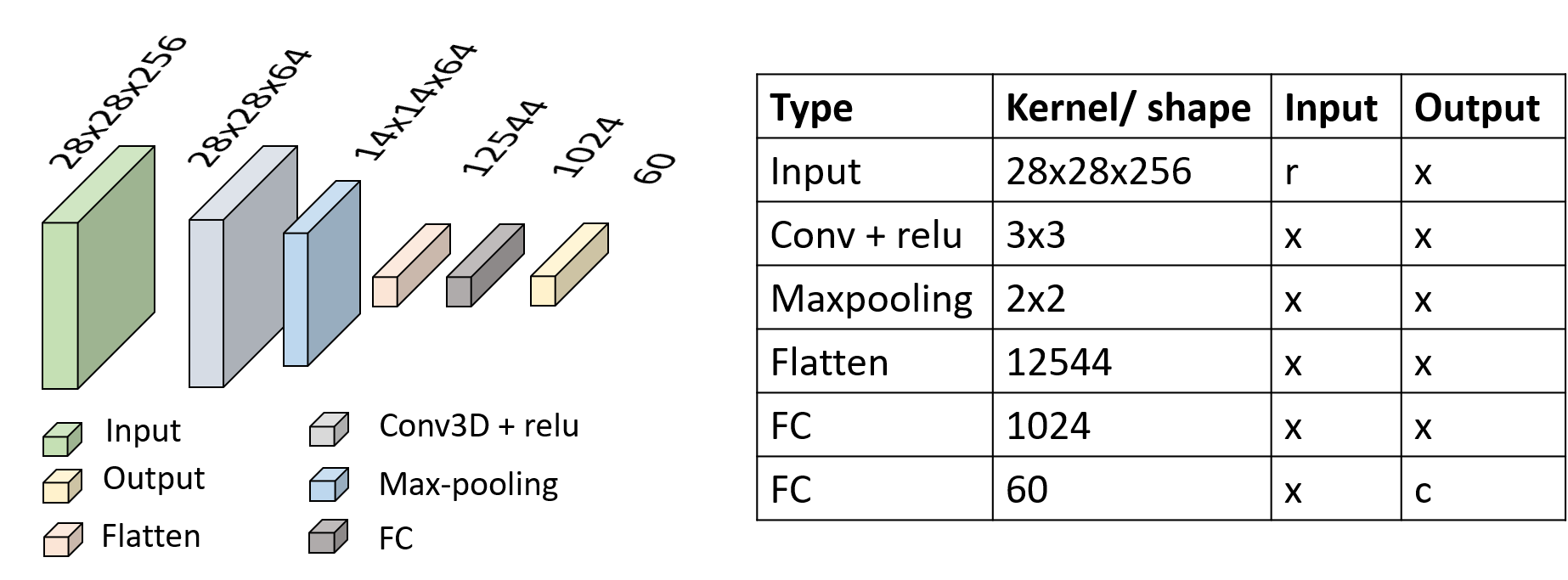}
		\end{center}
		\caption{Network architecture of the classification branch (CL-NET) of the multi-task approach. This network takes the learned representation $r$ as input and predicts class probabilities. FC: fully connected layer, c: predicted class probabilities.}
		\label{clnet}
	\end{figure*}
	
	\subsection{Video Rendering Network (VR-NET)}
	The video rendering network takes the learned representation $r$ along with query viewpoint $v$ and latent noise $z$ to synthesize a video clip. A detailed architecture of the proposed VR-NET is shown in Figure \ref{vrnet_3d}. It first integrates the representation $r$ with viewpoint $v$ and latent noise $z$ followed by convolution operation. A video is generated using convolutions along with upsampling of features. A similar network architecture is used for the experiments where images are generated using the learned representation. It uses 2D convolutions instead of 3D convolutions with no upsampling along the temporal axis in the feature space.

	\subsection{Classification Network (CL-NET)}
	
	We also propose to use the learned representation for activity classification. The network is trained for both video rendering as well as activity classification simultaneously in a multi-task learning. We added a classification branch (CL-NET) on top of representation which predicts class probabilities for activities. A detailed architecture for this branch is shown in Figure \ref{clnet}. It consists of 2D convolutions followed by fully connected layers for class predictions.
	

	\section{Experimental Details}
	
	\subsection{Hyperparameters}
	We implement the proposed method on Keras with Tensorflow backend. We train all our networks using Adam optimizer \cite{kingma2014adam} with a learning rate of 2e-5, $\beta1$=0.9, $\beta2$=0.999, and a decay of 1e-6. The networks were trained until convergence of loss. The batch normalization layers use a momentum of 0.9 with $\epsilon$=1e-3 and no scaling. The CL-NET branch has dropouts after every fully connected layer where we use dropout rate of 0.5. We use a skip rate of 3 frames for all our experiments.
	
	\subsection{Viewpoint}
\begin{figure}[ht]
		\begin{center}
			\includegraphics[width=0.5\linewidth]{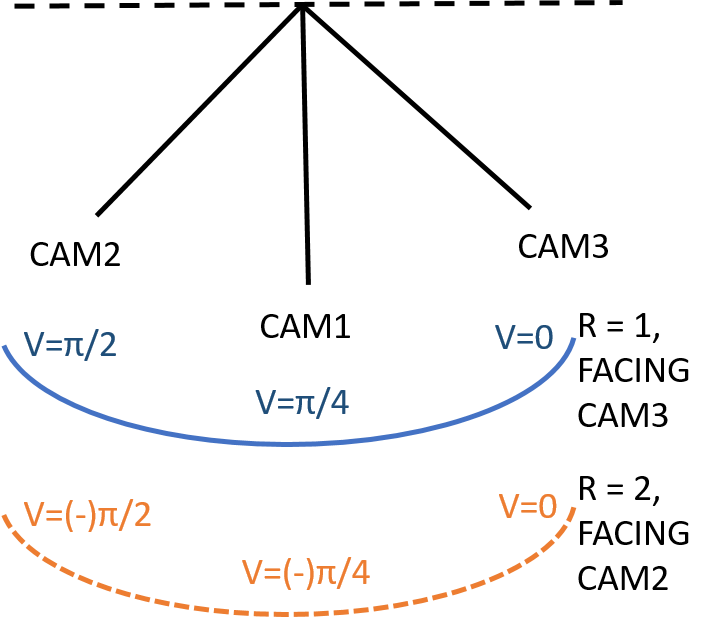}
		\end{center}
		\caption{\small Assimilation of viewpoint and actor orientation in NTU-RGB+D dataset. There are three different cameras (labeled as CAM1, CAM2 and CAM3) in the set-up. R  represents which camera the actor is facing: R=1 when actor faces CAM3 and R=2 when actor faces CAM2.}
		\label{fig2}
	\end{figure}

	The NTU-RGB+D dataset was captured using three cameras positions (CAM1, CAM2 and CAM3) and action videos were recorded in two conditions:  each actor facing CAM2, and each actor facing CAM3. The viewpoint involved 6 parameters including, camera height, camera distance, camera number, horizontal-pan, vertical-pan, and actor orientation. Horizontal and vertical pan in our training set was determined based on the cropping of the input frames (112, 112, 3) which was done randomly. The camera position was encoded based on its location using V values (-$\pi/2$, -$\pi/4$, 0, $\pi/4$, $\pi/2$) (Figure \ref{fig2}). The other viewpoint parameters were normalized between 0-1.
	
	\begin{table}[t!]
		\begin{center}
			\begin{tabular}{|l|p{25mm}|}
				\hline
				Method & Inception score \\
				\hline\hline
				\cite{vondrick2016generating} & 8.18$\pm$0.05  \\
				\hline
				\cite{saito2017temporal} & 11.85$\pm$0.07  \\
				\hline
				\cite{tulyakov2017mocogan} & 12.42$\pm$0.03  \\
				\hline
				Proposed & \textbf{17.18}$\pm$\textbf{0.19} \\
				\hline
				Original vidoes & 52.20$\pm$0.82 \\
				\hline
			\end{tabular}
		\end{center}
		\caption{Comparison of inception score with state-of-the-art approaches for video generation on UCF-101 dataset. Last row shows the inception score for the ground truth real videos.}
		\label{table_s1}
	\end{table}
	
	\subsection{PSNR and SSIM Evaluation}
	The quantitative evaluation presented in Table 1 and Table 2 of the manuscript was performed over 378 test videos from UCF-101 dataset for a fair comparison with existing methods. We compute the Peak Signal to Noise Ratio (PSNR) and the Structural Similarity Index Measure (SSIM) of \cite{wang2004image}. This experimental setup was suggested in \cite{mathieu2015deep} where every 10th video was selected for evaluation from the test list. Apart from this, the evaluation for Table 1 was performed only in the moving areas which was determined using optical flow \cite{revaud2015epicflow}. We use the optical flow images provided by \cite{mathieu2015deep} for our evaluation. In Table 2, the evaluation was performed over the full image for a fair comparison with \cite{mathieu2015deep}. The evaluation on NTU-RGB+D dataset was performed on all the testing videos where the target video was sampled randomly from each instance.
	
	\section{Additional Results}
	
	\subsection{Inception Score}
	
	We have recently seen some works for video synthesis which make use of adversarial loss in the GAN framework \cite{vondrick2016generating, saito2017temporal, tulyakov2017mocogan}. We compare the quality of the generated videos with these works in terms of inception score \cite{saito2017temporal}. We need a trained classifier to compute the inception score and we use a C3D model \cite{tran2015learning} pre-trained on Sports1M dataset \cite{karpathy2014large} and fine-tuned on UCF-101 as suggested in \cite{saito2017temporal}. We generated around 50K video samples for future prediction using the test set of UCF-101 with 6 frames each. A comparison is shown in Table \ref{table_s1} and we observe that the quality of videos generated using our proposed approach is much better as compared with these methods in terms of inception score.
	

	\begin{figure}[t!]
		\begin{center}
			\includegraphics[width=0.75\linewidth]{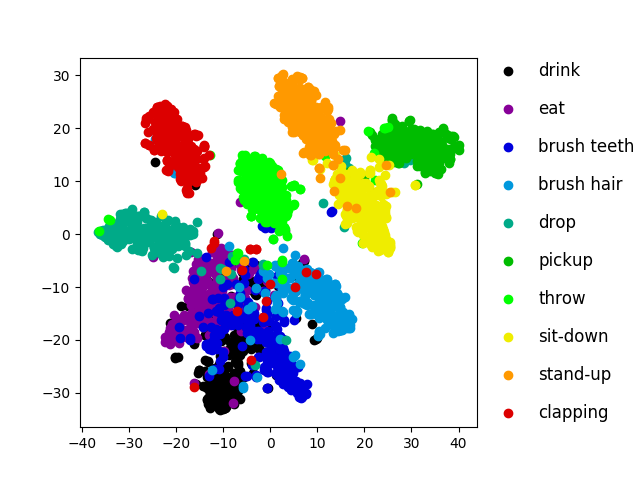}
		\end{center}
		\caption{t-SNE visualization of activity representations for a subset of 10 activities (out of 60) on NTU-RGB+D dataset. The representation is learned with \textit{one} input view to RL-NET. Most of the actions are well separated and the actions with similar appearance and dynamics are close to each other. The separation is not as good as when we use multiple views for representation learning (Figure 11, main manuscript) in case of confusing classes, however for others there is a clear visible boundary.}
		\label{fig_s12}
	\end{figure}
	
	\begin{figure}[t!]
		\begin{center}
			\includegraphics[width=0.95\linewidth]{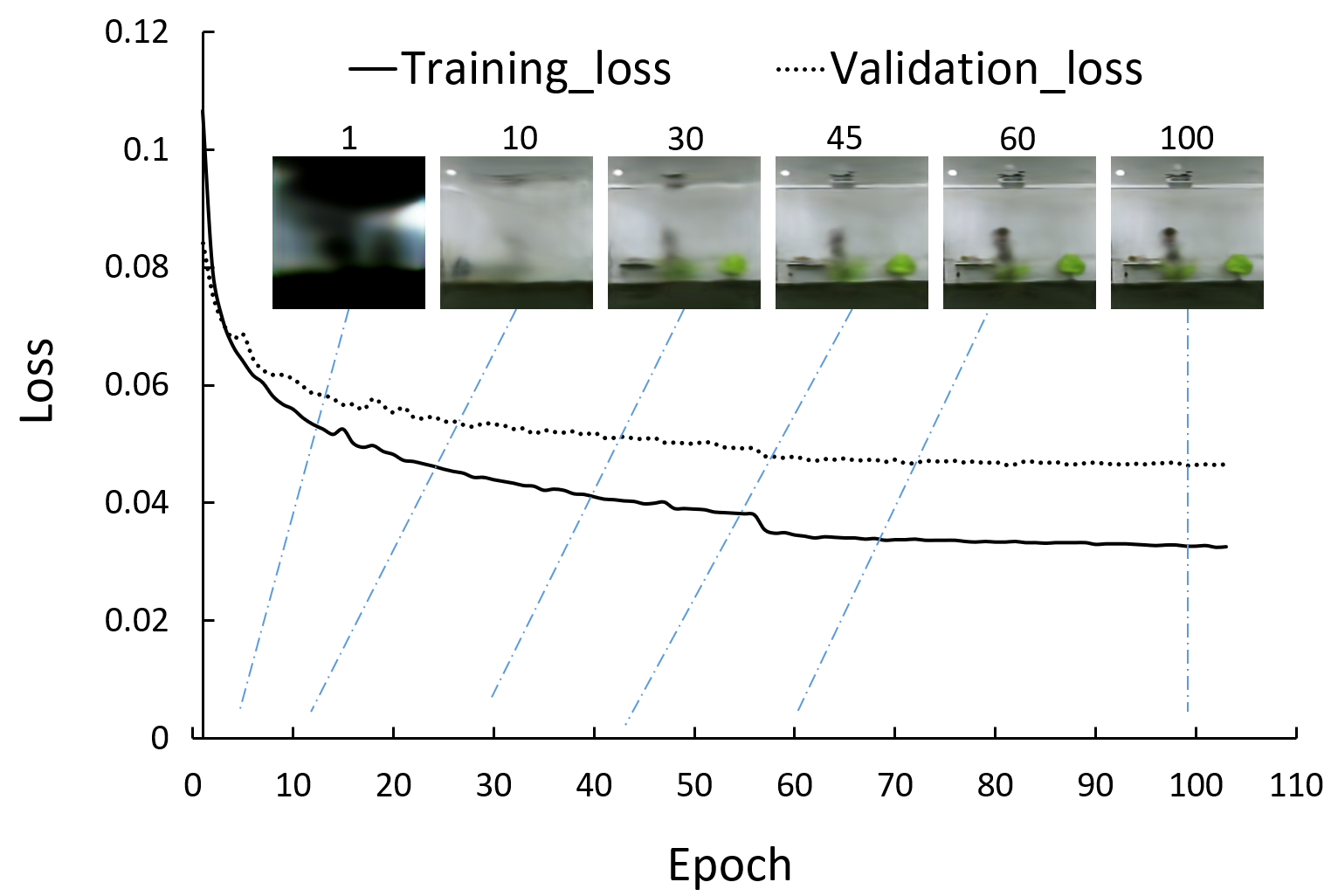}
		\end{center}
		\caption{The variation in training and validation loss as training progresses. The images are the first frame of the generated videos using the trained model at corresponding epoch.}
		\label{loss_unsupervised}
	\end{figure}

	
	\begin{figure*}[t!]
		\begin{center}
			\includegraphics[width=0.45\linewidth]{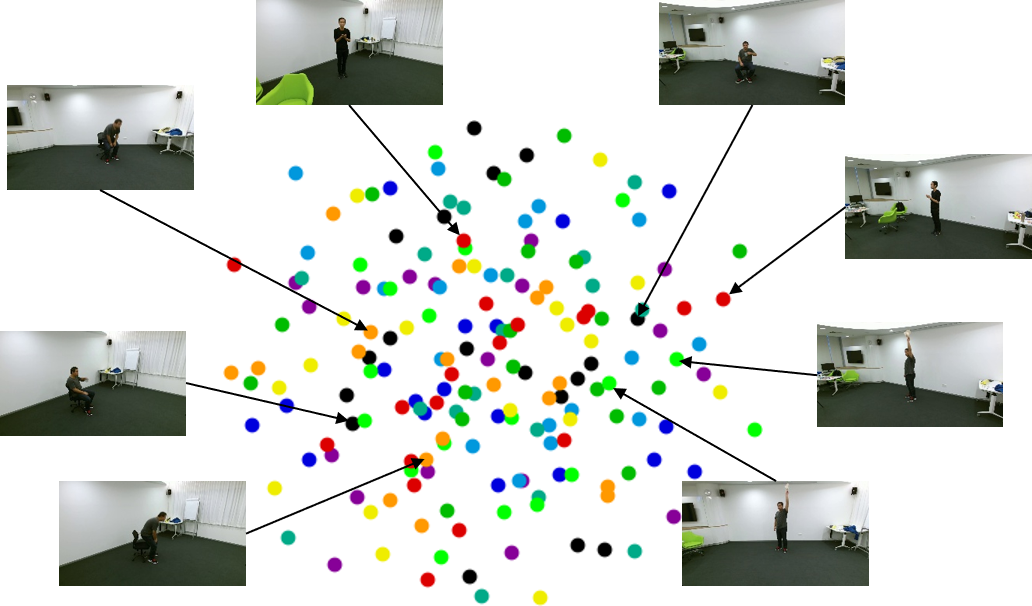} ~~~~~
			\includegraphics[width=0.45\linewidth]{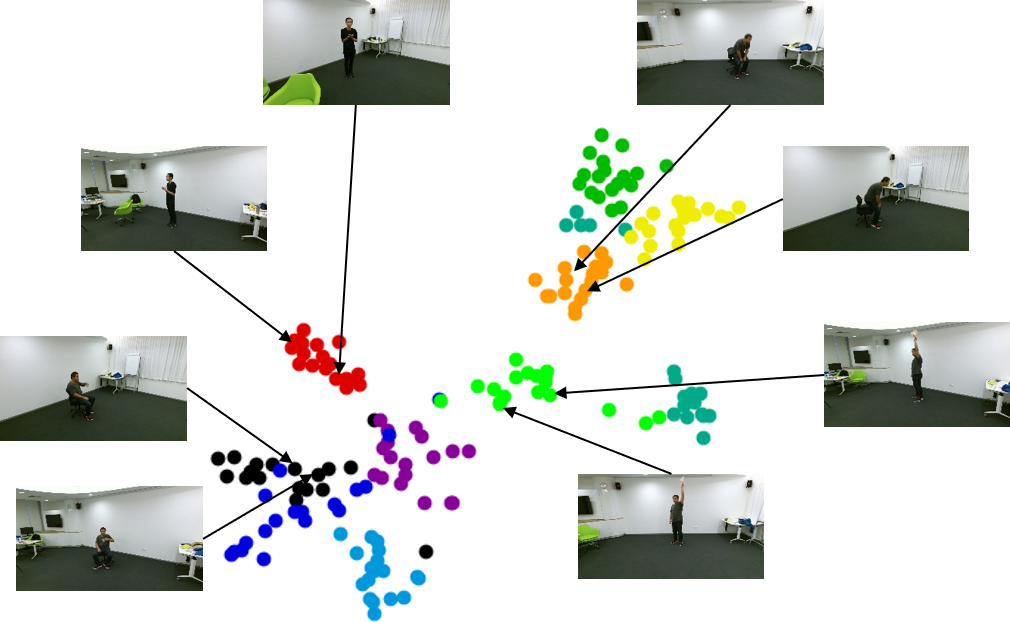}
		\end{center}
		\caption{View-invariant representation learning. t-SNE visualization of representations learned with VAE (left) and RL-NET (right) for a subset of 10 activities (out of 60, shown in Figure 6) on NTU-RGB+D dataset. The representation is inferred with \textit{one} input view to both VAE and RL-NET from only one scene (scene S001) in NTU-RGB+D dataset. The shown images are the first frame of the video clip. We observe that VAE is indifferent to view awareness of activities and mostly clusters videos with similar visual content. On the other hand, the proposed method is able to cluster activities from different views close to each other even if they have different viewpoints.}
		\label{fig_tsne_2}
	\end{figure*}

	\begin{figure*}[ht!]
		\centering
		\begin{subfigure}[b]{0.49\linewidth}
			\centering
			\includegraphics[width=0.15\linewidth]{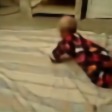}
			\includegraphics[width=0.15\linewidth]{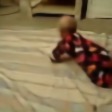}
			\includegraphics[width=0.15\linewidth]{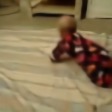}
			\includegraphics[width=0.15\linewidth]{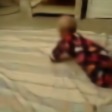}
			\includegraphics[width=0.15\linewidth]{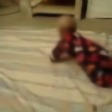}
			\includegraphics[width=0.15\linewidth]{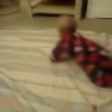} \\
			
			\includegraphics[width=0.15\linewidth]{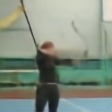}
			\includegraphics[width=0.15\linewidth]{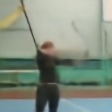}
			\includegraphics[width=0.15\linewidth]{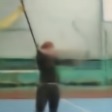}
			\includegraphics[width=0.15\linewidth]{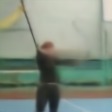}
			\includegraphics[width=0.15\linewidth]{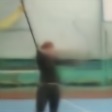}
			\includegraphics[width=0.15\linewidth]{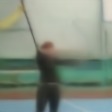} \\
			
			\includegraphics[width=0.15\linewidth]{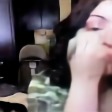}
			\includegraphics[width=0.15\linewidth]{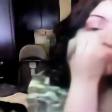}
			\includegraphics[width=0.15\linewidth]{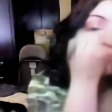}
			\includegraphics[width=0.15\linewidth]{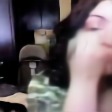}
			\includegraphics[width=0.15\linewidth]{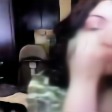}
			\includegraphics[width=0.15\linewidth]{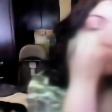} \\
			
			\includegraphics[width=0.15\linewidth]{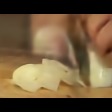}
			\includegraphics[width=0.15\linewidth]{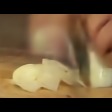}
			\includegraphics[width=0.15\linewidth]{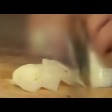}
			\includegraphics[width=0.15\linewidth]{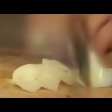}
			\includegraphics[width=0.15\linewidth]{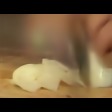}
			\includegraphics[width=0.15\linewidth]{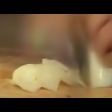} \\
			
			\includegraphics[width=0.15\linewidth]{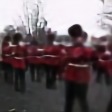}
			\includegraphics[width=0.15\linewidth]{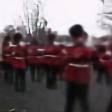}
			\includegraphics[width=0.15\linewidth]{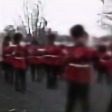}
			\includegraphics[width=0.15\linewidth]{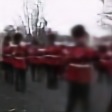}
			\includegraphics[width=0.15\linewidth]{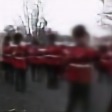}
			\includegraphics[width=0.15\linewidth]{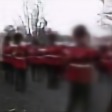}
			\caption{Generated video frames} 
			\label{fig_s1a} 
			
		\end{subfigure}
		\begin{subfigure}[b]{0.49\linewidth}
			\centering
			
			\includegraphics[width=0.15\linewidth]{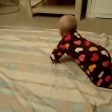}
			\includegraphics[width=0.15\linewidth]{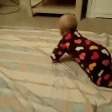}
			\includegraphics[width=0.15\linewidth]{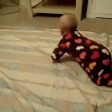}
			\includegraphics[width=0.15\linewidth]{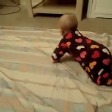}
			\includegraphics[width=0.15\linewidth]{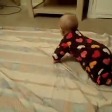}
			\includegraphics[width=0.15\linewidth]{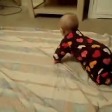}\\
			
			\includegraphics[width=0.15\linewidth]{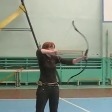}
			\includegraphics[width=0.15\linewidth]{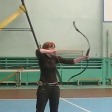}
			\includegraphics[width=0.15\linewidth]{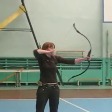}
			\includegraphics[width=0.15\linewidth]{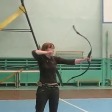}
			\includegraphics[width=0.15\linewidth]{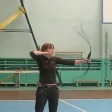}
			\includegraphics[width=0.15\linewidth]{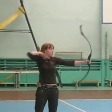}\\
			
			\includegraphics[width=0.15\linewidth]{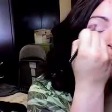}
			\includegraphics[width=0.15\linewidth]{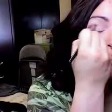}
			\includegraphics[width=0.15\linewidth]{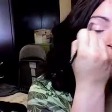}
			\includegraphics[width=0.15\linewidth]{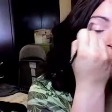}
			\includegraphics[width=0.15\linewidth]{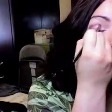}
			\includegraphics[width=0.15\linewidth]{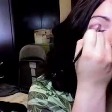}\\
			
			\includegraphics[width=0.15\linewidth]{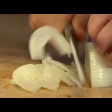}
			\includegraphics[width=0.15\linewidth]{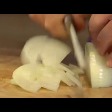}
			\includegraphics[width=0.15\linewidth]{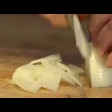}
			\includegraphics[width=0.15\linewidth]{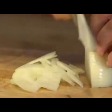}
			\includegraphics[width=0.15\linewidth]{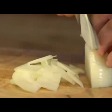}
			\includegraphics[width=0.15\linewidth]{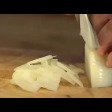}\\
			
			\includegraphics[width=0.15\linewidth]{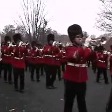}
			\includegraphics[width=0.15\linewidth]{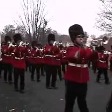}
			\includegraphics[width=0.15\linewidth]{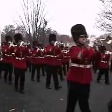}
			\includegraphics[width=0.15\linewidth]{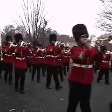}
			\includegraphics[width=0.15\linewidth]{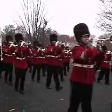}
			\includegraphics[width=0.15\linewidth]{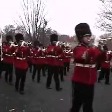}
			
			\caption{Ground-truth video frames} 
			\label{fig_s1b} 
			
		\end{subfigure}
		\caption{Future video frames generated (on Left) with 3D-Convolution Network (RL-NET and VR-NET) and their ground truth (on Right) on UCF-101 dataset. Row-1: Baby crawling, Row-2: Archery, Row-3: Apply eye make-up, Row-4: Cutting in kitchen, and Row-5: Band marching. 
		}
		\label{ucfFuture}
	\end{figure*}
	
	\subsection{Visualizing Embeddings}
	In the main manuscript, we discuss the 2-dimensional t-distributed stochastic neighbor embedding (t-SNE) \cite{maaten2008visualizing} visualization of the learned embeddings with two random views as inputs while generating the third view along with activity classification. Here we present the t-SNE visualization of the learned embedding from single views. The visualization of the embeddings from the final fully connected layer of CL-NET is shown in Figure \ref{fig_s12}. We observe that the separation for the activity classes which are confusing (such as drinking and eating) is slightly overlapping and it is not as sharp as we observe for multi-view embeddings. However, the network is still able to separate other activity classes very well.
	
	We also compare the representation learned by the proposed RL-NET with autoencoding density models such as Variational Autoencoder (VAE) \cite{kingma2013auto}. The VAE was implemented by replacing the RL-NET model with a CNN network (similar to BASENET) and keeping the rest of the network similar to ours. We observe that the proposed method was able to place the instances from similar classes close to each other despite the change in the viewpoint. VAE on the other hand failed to capture any structure in the representations with varying viewpoints and activity classes. A t-SNE comparison plot is shown in Figure \ref{fig_tsne_2} where we use samples only from one scene (scene S001) of NTU-RGB+D dataset.
	
	\subsection{Quality of Generated Videos}
	We also explore the variation in the quality of generated videos as the training progresses. The variation in the training and validation loss at each epoch is shown in Figure \ref{loss_unsupervised}. The graph also shows the first frame of the videos generated after each epoch with our trained model. The videos are generated with a 3D convolution RL-NET and VR-NET architecture, which was trained with input videos from two randomly selected views for each video and the third view was used for generation. We render the same view point (view 1) for the shown graph. We observe a gradual improvement in the visual quality of the generated video frames as the loss goes down during the training. The network first learns to synthesize the scene with the viewpoint, and later focuses on refining the actor and the corresponding action.
	
	
	\begin{figure}[t!]
		\centering
		
		\includegraphics[width=0.15\linewidth]{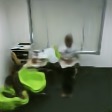}
		\includegraphics[width=0.15\linewidth]{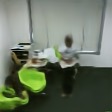}
		\includegraphics[width=0.15\linewidth]{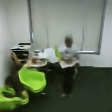}
		\includegraphics[width=0.15\linewidth]{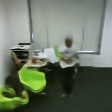}
		\includegraphics[width=0.15\linewidth]{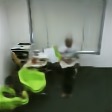}
		\includegraphics[width=0.15\linewidth]{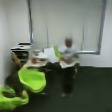}\\
		
		\includegraphics[width=0.15\linewidth]{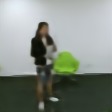}
		\includegraphics[width=0.15\linewidth]{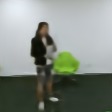}
		\includegraphics[width=0.15\linewidth]{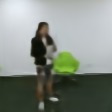}
		\includegraphics[width=0.15\linewidth]{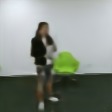}
		\includegraphics[width=0.15\linewidth]{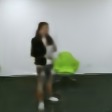}
		\includegraphics[width=0.15\linewidth]{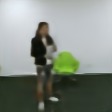}\\
		
		\includegraphics[width=0.15\linewidth]{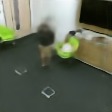}
		\includegraphics[width=0.15\linewidth]{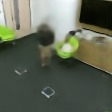}
		\includegraphics[width=0.15\linewidth]{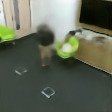}
		\includegraphics[width=0.15\linewidth]{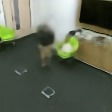}
		\includegraphics[width=0.15\linewidth]{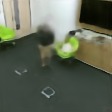}
		\includegraphics[width=0.15\linewidth]{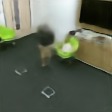}
		\caption{Video frames generated for arbitrary selected query time and view on NTU-RGB+D dataset. The network is given video clips from three views at randomly selected time. This network was trained on input video clips from all the three views and a video was generated from any one of these views at arbitrary time. view 3D-Convolution Network for action classes. Row-1: Stomachache/heart pain, Row-2: Eat meal/snack, and Row-3: Nod head/bow. 
		}
		\label{NTU_trained}
	\end{figure}
	
	\begin{figure}[t!]
		\centering
		
		\includegraphics[width=0.15\linewidth]{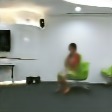}
		\includegraphics[width=0.15\linewidth]{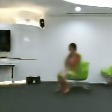}
		\includegraphics[width=0.15\linewidth]{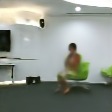}
		\includegraphics[width=0.15\linewidth]{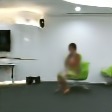}
		\includegraphics[width=0.15\linewidth]{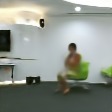}
		\includegraphics[width=0.15\linewidth]{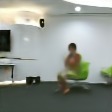}\\
		
		\includegraphics[width=0.15\linewidth]{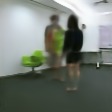}
		\includegraphics[width=0.15\linewidth]{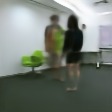}
		\includegraphics[width=0.15\linewidth]{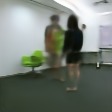}
		\includegraphics[width=0.15\linewidth]{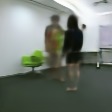}
		\includegraphics[width=0.15\linewidth]{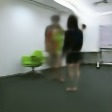}
		\includegraphics[width=0.15\linewidth]{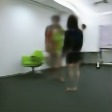}\\
		
		\includegraphics[width=0.15\linewidth]{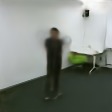}
		\includegraphics[width=0.15\linewidth]{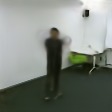}
		\includegraphics[width=0.15\linewidth]{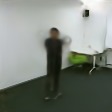}
		\includegraphics[width=0.15\linewidth]{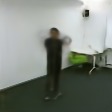}
		\includegraphics[width=0.15\linewidth]{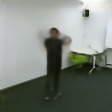}
		\includegraphics[width=0.15\linewidth]{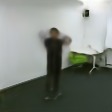}\\
		
		\includegraphics[width=0.15\linewidth]{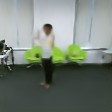}
		\includegraphics[width=0.15\linewidth]{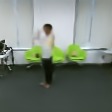}
		\includegraphics[width=0.15\linewidth]{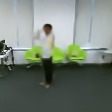}
		\includegraphics[width=0.15\linewidth]{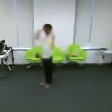}
		\includegraphics[width=0.15\linewidth]{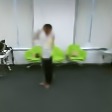}
		\includegraphics[width=0.15\linewidth]{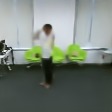}\\
		
		\includegraphics[width=0.15\linewidth]{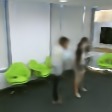}
		\includegraphics[width=0.15\linewidth]{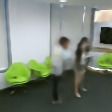}
		\includegraphics[width=0.15\linewidth]{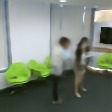}
		\includegraphics[width=0.15\linewidth]{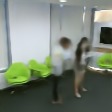}
		\includegraphics[width=0.15\linewidth]{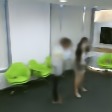}
		\includegraphics[width=0.15\linewidth]{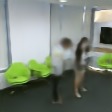}\\
		\caption{Video frames generated for future time for a given view on NTU-RGB+D dataset. Row-1: Wear on glasses, Row-2: Touch other person's pocket, Row-3: Brushing hair, Row-4: Brushing hair, and Row-5: Touch other person's pocket.}
		\label{NTU-prior}
	\end{figure}
	
	\subsection{Qualitative Results}
	Here we are showing some additional qualitative results for future frame generation on UCF-101 dataset and view and time aware video generation on NTU-RGB+D dataset. The generated frames are shown in Figure \ref{ucfFuture}, \ref{NTU_trained}, and \ref{NTU-prior}. We also provide the videos (gif animations) created using these frames in the supplementary material. The videos show the six frames generated by the model and their name corresponds to the activity class.

\end{document}